\documentclass[]{fairmeta}
\usepackage{enumitem}
\usepackage{float}
\usepackage{cuted}
\usepackage{subcaption}
\usepackage{soul}
\usepackage{multirow}
\usepackage{makecell}
\usepackage{amssymb}
\usepackage{tikz}
\newcommand{\tikzxmark}{%
\tikz[scale=0.23] {
    \draw[line width=0.7,line cap=round] (0,0) to [bend left=6] (1,1);
    \draw[line width=0.7,line cap=round] (0.2,0.95) to [bend right=3] (0.8,0.05);
}}
\newcommand{\tikzcmark}{%
\tikz[scale=0.23] {
    \draw[line width=0.7,line cap=round] (0.25,0) to [bend left=10] (1,1);
    \draw[line width=0.8,line cap=round] (0,0.35) to [bend right=1] (0.23,0);
}}

\newcommand\sdpabsone{1.07$\times$}
\newcommand\sdpabsmax{1.43$\times$}
\newcommand\compileoversdpabsone{2.14$\times$}
\newcommand\compileoversdpabsmax{2.16$\times$}
\newcommand\compileovervanillabsone{2.28$\times$}
\newcommand\compileovervanillabsmax{3.09$\times$}
\newcommand\autoquantovervanillabsone{2.13$\times$}
\newcommand\autoquantovervanillabsmax{4.38$\times$}
\newcommand\autoquantoversdpabsmax{3.06$\times$}
\newcommand\autoquantovercompilebsone{1.20$\times$}
\newcommand\autoquantovercompilebsmax{1.57$\times$}

\newcommand\layerskipspeedup{1.58$\times$}
\newcommand\layerskipsystemspeedup{3.88$\times$}

\newcommand\layerskipcodellamaseven{1.59$\times$}
\newcommand\layerskipcodellamathirty{1.53$\times$}

\newcommand\layerskipchameleonmscoco{1.43$\times$}
\newcommand\layerskipchameleonvizwiz{1.83$\times$}

\title{Characterizing and Efficiently Accelerating Multimodal Generation Model Inference}

\author{Yejin Lee}
\author{Alicia Golden}
\author{Anna Sun}
\author{Basil Hosmer}
\author{Bilge Acun}
\author{Can Balioglu}
\author{Changhan Wang}
\author{Charles David Hernandez}
\author{Christian Puhrsch}
\author{Daniel Haziza}
\author{Driss Guessous}
\author{Francisco Massa}
\author{Jacob Kahn}
\author{Jeffrey Wan}
\author{Jeremy Reizenstein}
\author{Jiaqi Zhai}
\author{Joe Isaacson}
\author{Joel Schlosser}
\author{Juan Pino}
\author{Kaushik Ram Sadagopan}
\author{Leonid Shamis}
\author{Linjian Ma}
\author{Min-Jae Hwang}
\author{Mingda Chen}
\author{Mostafa Elhoushi}
\author{Pedro Rodriguez}
\author{Ram Pasunuru}
\author{Samuel Hsia}
\author{Scott Yih}
\author{Sravya Popuri}
\author{Xing Liu}
\author{Carole-Jean Wu}

\affiliation{AI Research at Meta}

\abstract{Generative artificial intelligence (AI) technology is revolutionizing the computing industry, posing new system design and optimization opportunities. In particular, AI's ability to understand and respond in multiple modalities comes with significant system resource demands. To sustainably scale generative AI capabilities to billions of users in the world, inference must be fast and efficient. This paper pinpoints key system design and optimization opportunities by characterizing a family of emerging multi-modal generation models on real systems. Auto-regressive token generation is a critical latency performance bottleneck, typically dominated by GPU idle time. In addition to memory-intensive attention across the generative AI models, linear operations constitute significant inference latency due to the feed forward networks in Transformer-based models. We demonstrate that state-of-the-art optimization levers, spanning from applications to system software and hardware, set a 
\layerskipsystemspeedup~better baseline.}

\date{\today}
\correspondence{Yejin Lee \email{yejinlee@meta.com}, Carole-Jean Wu \email{carolejeanwu@meta.com}}

\begin{document}

\maketitle

\section{Introduction}
\label{sec:introduction}

% \all
Generative AI technologies are driving an unprecedented growth for the computing industry, introducing a new paradigm shift for AI. This technology redefines the interaction between humans and AI by enabling the creation of highly realistic images~\cite{emuedit}, videos~\cite{makeavideo,emuvideo}, texts, and speech~\cite{seamless}, as well as intricate textual patterns or even new materials. 
Large language models (LLMs), such as ChatGPT~\cite{chatgpt}, Llama~\cite{llama,llama2}, or Gemini~\cite{gemini}, demonstrate remarkable capabilities. LLMs not only enhance user experience by providing contextually relevant interactions but also play a critical role in automating complex tasks. It has already germinated a wide variety of applications, leading to higher productivity.

Beyond LLMs, multi-lingual speech translation and transcription models, such as Seamless~\cite{seamless}, Whisper~\cite{whisper}, or Translatotron~\cite{translatotron}, are pivotal in breaking down the language barriers and enhancing communication on a global scale. The speech models provide accurate and real-time translation/transcription across different languages by processing speech and text modalities together, such as Speech to Speech and Text (S-ST), Text to Speech and Text (T-ST), and Automatic Speech Recognition (ASR).

In addition to text and speech modalities, state-of-the-art AI technologies can take inputs of multiple modalities to serve multi-modal use cases. Taking Chameleon~\cite{chameleon} as an example, this multi-modal foundation model can take images and text as input and generate outputs in either modality. Such models are the foundation of image editing or visual question-answer (VQA) use cases. Also, the multi-modal models are capable of image generation based on text prompts or even ChatBot style conversations.

\begin{figure}
  \centering
  \includegraphics[width=0.65\columnwidth]{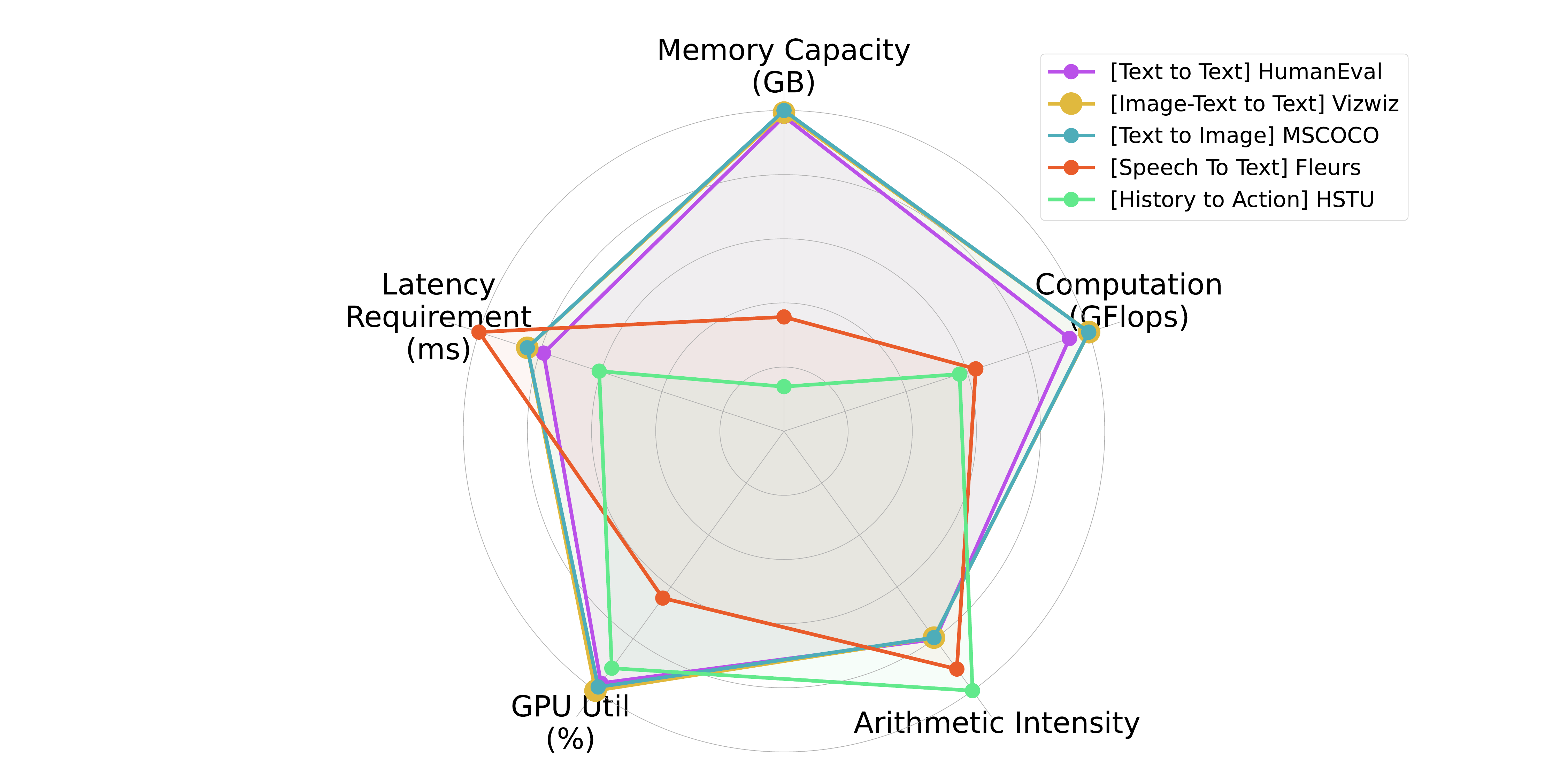}
  % \vspace{-0.1in}
  \caption{Multi-modal generation tasks exhibit distinct system requirements  across end-to-end inference latency, GPU utilization, memory capacity and computation requirement.}
  \label{fig:motivation}
  \vspace{-0.2in}
\end{figure}

Beyond learning from texts, language, speeches, images or videos, generative AI technologies are also adopted in deep learning recommendation systems as well. Leveraging the ability of Attention-based Transformers for automatically extracting and learning features from datasets, recent deep learning recommendation models, such as HSTU~\cite{hstu}, TIGER~\cite{tiger}, introduce a new feature generation paradigm by adopting sequential generative models. Such new model architecture uses generative models to accurately predict  items of interest. Generative recommendation models overcome the model quality saturation problem faced by existing deep learning recommendation models (DLRMs)~\cite{dlrm}, exceeding prediction quality over prior recommender system technologies.

While a disproportional investment is currently focused on LLMs, generative AI technologies that are capable of processing multi-modal inputs and outputs are on the horizon. Depending on distributions of input prompt lengths and use cases  (Section~\ref{sec:seqlen}) and the characteristics of model architectures (Section~\ref{sec:opportunity}), the system design space for efficiency presents unique optimization opportunities. For example, a recent work shows that training a state-of-the-art text-to-image model can use \textit{14x more GPUs per model parameter} than that of an industry-scale LLM~\cite{golden2023generative}. To efficiently accelerate multi-modal generation model inference, this paper provides an in-depth system performance characterization for important industry-scale generative AI tasks: language (Code Llama~\cite{codellama}), speech translation (Seamless~\cite{seamless}),  text and image generation (Chameleon~\cite{chameleon}), and generative deep learning recommender systems (gDLRM~\cite{hstu}).

These models serve important roles in Meta's workloads, with Llama functioning as our foundational large language model powering Meta AI's core capabilities. Seamless delivers high-quality translation services across Meta platforms, enabling multilingual access to Instagram and Facebook content. Chameleon serves as a foundation model for multimodal generation, handling various combinations of text and image inputs/outputs, while HSTU helps drive recommendation systems, processing billions of recommendations daily with strict latency constraints.

\begin{figure*}
  \begin{subfigure}{.23\textwidth}
  \centering
    \includegraphics[page=2,width=1\linewidth]{analysis_figures_revision/hpca_figure.pdf}
    \caption{Code Llama}
    \label{fig:codallamaarch}
  \end{subfigure}
  \hfill
  \begin{subfigure}{.23\textwidth}
  \centering
    \includegraphics[page=3,width=1\linewidth]{analysis_figures_revision/hpca_figure.pdf}
    \caption{Chameleon}
    \label{fig:chameleonarch}
  \end{subfigure}
  \hfill
  \begin{subfigure}{.23\textwidth}
  \centering
    \includegraphics[page=4,width=1\linewidth]{analysis_figures_revision/hpca_figure.pdf}
    \caption{Seamless}
    \label{fig:seamlessarch}
  \end{subfigure}
  \hfill
  \begin{subfigure}{.23\textwidth}
  \centering
    \includegraphics[page=1,width=1\linewidth]{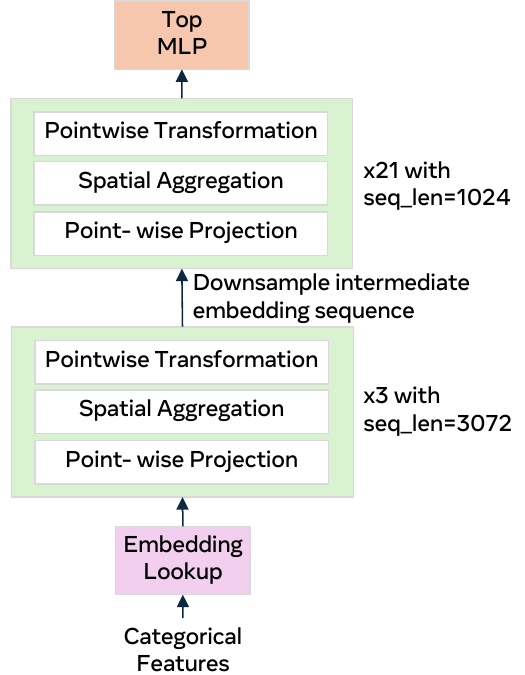}
    \caption{HSTU}
    \label{fig:hstuarch}
  \end{subfigure}
  \hfill
  \caption{Model Architectures of Llama~\cite{codellama}, Seamless~\cite{seamless}, Chameleon~\cite{chameleon}, HSTU~\cite{hstu}.}
  \label{fig:genai-model-architectures}
\end{figure*}

To sustainably scale generative AI technologies for a large, diverse variety of applications~\cite{wu2024sustainable}, we must understand and enable AI deployment in a resource-efficient manner~\cite{wu2022sustainable}. 
Figure~\ref{fig:motivation} illustrates the system requirements for four multi-modal generation models at a single batch size. The chart highlights the latency requirement, overall memory capacity, communication requirement, and GPU utilization,  for different tasks across these models. It is evident that, depending on \textit{input modalities and model architectures of a specific task}, system resource utilization characteristics are distinct. For example, Chameleon can perform image-to-text (I-T), text-to-image (T-I), and image-text-to-text (IT-T) without requiring fine-tuning for each task. However, the T-I task demands significantly higher resource requirements across all four axes.

To scale advanced generative AI capabilities to billions of users in the world, inference needs to complete on the order of milliseconds and efficiently. 
The in-depth real-system performance characterization results in Section~\ref{sec:opportunity} guide the focus of inference performance and efficiency optimization. We take a step further to enable state-of-the-art inference performance optimization techniques --- torch.compile and CUDA Graph for memory efficiency optimization~\cite{pytorch-2.0}, Scaled Dot Product Attention (SDPA) / Flash Attention to speed up Transformer's key performance bottleneck~\cite{flashattention}, quantization to further improve compute density and memory bandwidth utilization. When enabling state-of-the-art optimization levers properly, the inference performance over the important generative AI tasks can be improved by 3.8x, setting a new, more rigorous baseline. 
Beyond efficiently accelerating inference performance horizontally across the key generative AI tasks, in Section~\ref{sec:worklaodspecific}, we present ways to further improve inference performance efficiency with application-specific, algorithmic optimization. We enable LayerSkip~\cite{layerskip}, a self-speculative decoding approach to our workloads to speedup generation and show the inference performance is improved by \layerskipspeedup.
% sdpa_speedup:  0.9281622818369191 16.083346557692035
% torch_compile:  0 2.826836238460463
% torch_compile_autoquant:  0 2.830280988904695
% sdpa_speedup:  0.9732423126140957 11.915163412357675
% torch_compile:  0 1.3091661108395527
% torch_compile_autoquant:  0 1.631425810302218
The key contributions of this paper are as follows:

% \vspace{-0.175in}
\begin{itemize}[leftmargin=*]
    \item \textbf{System Performance Characterization for Emerging Multi-Modal Generative AI Tasks} This paper delivers an in-depth examination of system performance across four pivotal generative AI models: LLM (Code Llama), Speech Translation (Seamless), Generative Text and Image Models (Chameleon), and Generative Deep Learning Recommendation Models (gDLRM). Our analysis covers critical aspects, such as computational and memory bandwidth requirements, variations in input distributions and roofline analysis --- key to inference performance efficiency optimization.
    
    \item \textbf{Optimized Baseline for Generative AI Inference Acceleration} We demonstrate the importance of enabling state-of-the-art optimization methods --- torch.compile, CUDA Graph, SDPA/Flash Attention, and quantization --- that accelerate the inference performance across the generative AI tasks by upto 28$\times$. Algorithmic optimization --- LayerSkip --- improves inference performance as well by \layerskipspeedup. Altogether, cross-stack solutions, spanning algorithm and systems, improve inference performance by an average of \layerskipsystemspeedup.
    We also highlight the performance impact of using a newer generation of GPUs by comparing the performance analysis across different GPU generation.
    \item \textbf{Design Implications and New Directions for Future Systems} We distill the implications of our findings for future research and development --- \textit{1) New solutions must improve upon stronger baseline} \textit{2) With proper understandings of the distinct characteristics and end-to-end inference pipeline of a given model, we can achieve \layerskipsystemspeedup\ speedup with state-of-the-art optimizations leverages} \textit{3) Enhancing the baseline with software optimization methods unlocks new possibilities for current and future hardware architectures.}

\end{itemize}

% \vspace{-0.12in}
\section{Background and Motivation}

\subsection{Understanding the Lay of the Land for Multi-modal Generative AI Tasks}

We provide an overview for key generative AI technologies. 
Figure~\ref{fig:genai-model-architectures} illustrates the model architectures for four generative AI models --- LLM (Code Llama), Speech Translation (Seamless), Generative Text and Image Models (Chameleon) and Generative Deep Learning Recommendation Models (gDLRM). 
Table~\ref{tbl:workloads} summarizes input/output modalities and sequence length distribution for  different workloads.

\begin{table*}[]
\centering
\small
\begin{tabular}{l|l|c|l|l|ll}
\multirow{2}{*}{Category}                 & \multirow{2}{*}{Model} & \multirow{2}{*}{\makecell[l]{Auto-\\regressive}} & \multirow{2}{*}{\makecell[l]{Nota\\tion}} & \multirow{2}{*}{Tasks}                 & \multicolumn{2}{c}{Modality}                  \\ \cline{6-7}
                                          &                        &                           & &                                         & \multicolumn{1}{c|}{Input}          & \multicolumn{1}{c}{Output}   \\ \hline\hline
\makecell[l]{Text-based\\LLM}              & Llama                      & \tikzcmark & {\bf T-T}  & \makecell[l]{Code Completion,\\Infilling, Instruction} & \multicolumn{1}{l|}{Text}           & Text     \\ \hline
\multirow{3}{*}{\makecell[l]{Image\&Text\\Generation}}   & \multirow{3}{*}{Chameleon} & \multirow{3}{*}{\tikzcmark} & {\bf I-T}  & Image Captioning                                  & \multicolumn{1}{l|}{Image}          & Text     \\ \cline{4-6}
                                          &                            & & {\bf T-I}  & Image Generation                                  & \multicolumn{1}{l|}{Text}           & Image    \\ \cline{4-6}
                                          &                            & & {\bf IT-T} & Visual Question Ans.                         & \multicolumn{1}{l|}{Image \& Text}  & Text     \\ \hline
                                          % &                            & {\bf IT-I} & Text-based Image Editing                          & \multicolumn{1}{l|}{Image \& Text}  & Image    \\ \hline
\multirow{4}{*}{\makecell[l]{Speech\&Text\\Translation}} & \multirow{4}{*}{Seamless}  & \multirow{4}{*}{\makecell[c]{$\bigtriangleup$\\Only text\\decoder}} & {\bf S-S}  & Speech-to-Speech Trans.                      & \multicolumn{1}{l|}{Speech}         & Speech   \\ \cline{4-6}
                                          &                           & & {\bf S-T}  & Speech-to-Text Trans.                        & \multicolumn{1}{l|}{Speech}         & Text     \\ \cline{4-6}
                                          &                           & & {\bf T-T}  & Text-to-Text Trans.                          & \multicolumn{1}{l|}{Text}           & Text     \\ \cline{4-6}
                                          &                           & & {\bf T-S}  & Text-to-Speech Trans.                        & \multicolumn{1}{l|}{Text}           & Speech   \\ \hline
%old: Generative DLRM                           & HSTU~\cite{hstu}                      & {\bf ?} & Feature Generation                                & \multicolumn{1}{l|}{\makecell[l]{Sequentialized \\Categorical Feature}} & Feature \\ \hline

% User History alternatives: Sequentialized Categorical Features
\makecell[l]{Generative\\DLRM}                           & HSTU                      & \tikzxmark & {\bf H-A} & Ranking and Retrieval                     & \multicolumn{1}{l|}{User History} & \makecell[l]{Engagement Type\\(ranking)\\Recommend Item\\(retrieval)} \\ \hline
\end{tabular}
\caption{The input and output modality of each task performed by four multimodal generative models, LLM (Llama), speech\&text translation (Seamless), text\&image generation (Chameleon) and generative DLRM (HSTU).}
\label{tbl:workloads}
% \vspace{-0.3in}
\end{table*}

% \vspace{-0.1in}
\subsubsection{Llama for Language Generation}
% Code Llama is a family of large language models based on Llama-2 for coding tasks. 
% Code Llama takes text-based code as input and generates code as output. The input sequence length distribution for Code Llama can vary depending on the specific model and task it is being used for. However, in general, Code Llama models are trained on a wide range of input sequence lengths to be able to handle varying sizes of code snippets. For example, Code Llama support sequence lengths up to 100,000 tokens which is enough to capture a reasonably sized code snippet while keeping the computation feasible.

Code Llama is a large language model for coding tasks. Code Llama models are trained on a wide range of input sequence lengths to be able to handle varying sizes of code snippets. For example, Code Llama support sequence lengths up to 100,000 tokens which is enough to capture a reasonably sized code snippet while keeping the computation feasible.

Code Llama has a standard transformer architecture~\cite{attentionisallyouneed} as shown in Figure~\ref{fig:codallamaarch}. 
In this paper, we take Code Llama as our model representing language generative AI model. And we refer Code Llama as Llama from now on for convenience.
The model consists of embedding layer followed by consecutive Transformer decoder blocks that includes attention and feed forward layer. Specifically, Llama 34B model has 48 layers of Transformer decoder blocks.

% Llama is an autoregressive generation model where inference pipeline is broken down into two phases: prefill (\textbf{Prefill}) and incremental decoding (\textbf{Decoding}). Prefill takes the full input prompt whereas Decoding generates output tokens one by one based on previously generated tokens. In Decoding, KV cache optimization is key to relieve computational intensity.
% %overhead making computation to happen only for a single token that is generated in the previous incremental decoding step. 
% The input sequence length for Decoding is always 1 with KV cache optimization while the input sequence length for Prefill is determined by the length of the full input prompt.

% Llama is an autoregressive generation model where inference pipeline is broken down into two phases: prefill (\textbf{Prefill}) and incremental decoding (\textbf{Decoding}). 
% Prefill processes the entire input prompt at once, computing attention across all input tokens ($O(N^2)$ complexity), while Decoding generates output tokens sequentially with lower computational intensity but higher memory bandwidth requirements due to KV cache operations. In Decoding, KV cache optimization stores and reuses previously computed key-value pairs, making the input sequence length always 1, while the Prefill sequence length depends on the full input prompt.

Llama is an autoregressive generation model where inference pipeline is broken down into two phases: prefill (\textbf{Prefill}) and incremental decoding (\textbf{Decoding}). 
Prefill processes the entire input prompt of length $N$ at once, computing attention across all input tokens ($O(N^2)$ complexity), whereas Decoding generates output tokens one by one based on previously generated tokens. 
These phases present different computational characteristics: Prefill has high computational intensity due to processing full input sequence length $N$ while Decoding is computationally lighter by using KV cache optimization that stores and reuses key-value pairs, though this makes Decoding memory-bound due to frequent cache access.

\subsubsection{Chameleon for Text and Image Generation}\label{sec:chameleonbackground}
Chameleon is a foundational model for the family of early-fusion token-based mixed-modal models capable of understanding and generating images and text. It is capable of performing broad range of tasks including visual question answering, image captioning, text and image generation, and long-form mixed modal generation in a single model. The model architecture of Chameleon largely follows Llama-2~\cite{llama2} as shown in Figure~\ref{fig:chameleonarch}, thus Chameleon is also an auto-regressive generation model. For normalization, Chameleon continue to use RMSNorm~\cite{rmsnorm}; and Chameleon uses the SwiGLU~\cite{swiglu} activation function and rotary positional embeddings (RoPE)~\cite{rope}.
% Chameleon 34B model has 48 layers of Transformer decoder blocks.

Chameleon represents images, text, and code modalities as discrete tokens and uses a uniform transformer-based architecture that is trained from scratch in an end-to-end fashion on around 10T tokens of interleaved mixed-modal data. Chameleon can take any combination of image and text and utilizes image tokenizer~\cite{cm3imagetokenizer} and text tokenizer~\cite{cm3texttokenizer} respectively to generate tokens to be fed to the model. 
For text generation, generated tokens are decoded by text tokenizer to generate readable texts.
For image generation, Chameleon generates 1024 image tokens and then detokenize them using an image detokenizer to generate images in a format that can be interpreted by human such as jpg. And Chameleon uses a contrastive decoding method specifically for T-I task, which aims to maximize the differences between a weak and a strong model. 
Logits from conditioned outputs are treated as the strong model, while unconditional logits are considered as the weak model. 
As a result, Chameleon decodes twice at each time step for T-I task.

% \seamless
\subsubsection{Seamless for Speech Translation}
Seamless~\cite{seamless} is a family of speech translation models that enable more natural and authentic communication across languages. SeamlessM4T is the foundation model for multilingual multimodal machine translation supporting around 100 languages. SeamlessM4T achieves state-of-the-art semantic accuracy, supports a wide range of languages, and provides multitasking capabilities from and into text or speech. 
% [Carole] Cutting the introduction of Seamless Expressive and SeamlessStreaming since the rest of the paper does not use them at all.
%SeamlessM4T serves as foundation for SeamlessExpressive, a model that preserves elements of prosody and voice style across languages and SeamlessStreaming, a model supporting simultaneous translation and streaming ASR for around 100 languages. 
% SeamlessExpressive and SeamlessStreaming are combined into Seamless, a unified model featuring multilinguality, real-time and expressive translations. 
% In this paper, we focus on SeamlessM4T since it plays a role as a base model for other models in the family. From now on, we refer to SeamlessM4T as Seamless.

SeamlessM4T, which we refer to as Seamless in this paper, consists of multiple pretrained blocks that are finetuned as a unified model. The four main building blocks are shown in Figure~\ref{fig:seamlessarch},
\begin{itemize}[leftmargin=*]
    \item \textbf{Conformer Speech Encoder} (Blue) A speech representation learning model that leverages unlabeled speech audio data.
    \item \textbf{Text-to-Text Translator (T2TT)} (Pink) A text-to-text translation model pre-trained on NLLB data in nearly 100 languages. It is "\textit{only}" autoregressive module among all modules in Seamless.
    \item \textbf{Non-autoregressive (NAR) T2U} (Green) NAR T2U is a text-to-unit sequence-to-sequence module.
    \item \textbf{Vocoder} (Orange) A HiFi-GAN unit-vocoder that converts generated units to waveform output where an unit represents speech combining different aspects such as phonemes and syllables %, which can be used to generate sounds that are audible to humans. 
\end{itemize}

Seamless utilizes different set of modules according to the task it is performing.
For text generation tasks, such as S-T and T-T,the conformer speech encoder and T2TT modules are utilized. For speech generation tasks, such as T-S and S-S, NAR T2U and Vocoder are additionally activated and the output of the translated text from T2TT is fed as an input to NAR T2U.

% The inputs for SeamlessM4T is generated by extracting 80-dimensional filterbank features from the raw audio waveform at 100Hz frame rate and stacking every 2 frames for the final 160-dimensional 50Hz features. These extracted features (i.e., dimension 160 for Seamless M4T) become the input and the number of features becomes the sequence length of the model. 

\begin{table*}[]
\centering
\small
\begin{tabular}{l|l|l|l|l|l|l|l|l|l|l|l}
% \hline
\multirow{3}{*}{Model} & \multirow{3}{*}{Dataset} & \multicolumn{8}{c|}{Modality}                                                                                         & \multirow{3}{*}{\makecell[l]{Decode\\Step\\Count}} & \multirow{3}{*}{\makecell[l]{Avg.\\Time\\(ms)}} \\ \cline{3-10}
                       &                          & \multicolumn{4}{c|}{Input}                                                      & \multicolumn{4}{c|}{Output}         &           &                           \\\cline{3-10}
                       &                          & \multicolumn{1}{c|}{Modality} & \multicolumn{1}{c|}{Min} & \multicolumn{1}{c|}{Max} & \multicolumn{1}{c|}{Avg} & \multicolumn{1}{c|}{Modality} & \multicolumn{1}{c|}{Min} & \multicolumn{1}{c|}{Max} & \multicolumn{1}{c|}{Avg} & &  \\ \hline\hline
\multirow{2}{*}{Llama} & HumanEval & Text & 44 & 430 & 154    & Text    & 55    & 10000    &  692                        & 538   &    4494             \\ \cline{2-11}
                           & MBPP & Text    & 29 & 1748    & 59    & Text    & 38    & 10000    &  1076                        & 1016  & 5567              \\ \hline
\multirow{4}{*}{\makecell[l]{Seam-\\less}} & \multirow{4}{*}{\makecell[l]{Fleurs\\Eng-Spa}} & \multirow{2}{*}{Speech}    & \multirow{2}{*}{179} & \multirow{2}{*}{1464}    & \multirow{2}{*}{493}   & Speech    & 129    & 1029    &    385                      & \multirow{2}{*}{35}    &  1578           \\ \cline{7-10}
 & & & & & & Text    &  15   &  98   &       36                   &          & 1321   \\ \cline{3-11}
 & & \multirow{2}{*}{Text}    & \multirow{2}{*}{12} & \multirow{2}{*}{80}    & \multirow{2}{*}{31}   & Speech    & 145    & 1030    & 393                         & \multirow{2}{*}{34}       & 1432          \\ \cline{7-10}
 & & & & & & Text    &  14   & 95    &       35                  &    & 1187       \\ \hline
\multirow{4}{*}{\makecell[l]{Chame-\\leon}} & MSCOCO & Image    & 1030  & 1030    & 1030  & Text    & 30    & 30    &  30                        & 30    & 2913             \\ \cline{2-11}
                           & Vizwiz & Img\&Txt    & 1033  & 1095    & 1040  & Text    & 10    & 10    &  10                        & 10      & 1253         \\ \cline{2-11}
                           & \makecell[l]{MSCOCO} & Text    & 10  & 22    & 13.9  & Image    & O(1025)    & O(1025) & O(1025)  & 1024 & 159702  \\ \hline
HSTU & Synthetic & \makecell[l]{User\\History}    & 4507 & 5121 & 4814 & Action    & 4507.0 & 5121.0 & 4813.9 & N/A & 50       \\ \hline

\end{tabular}
\caption{Sequence Length Distribution of Four Generative AI Models. We use 5 sample for each workload.}
\label{tbl:seq_len}
\vspace{-0.35cm}
\end{table*}

% \color{black}

% this is the original version (as of 06/15)
%\subsubsection{gDLRM for Recommender System Feature Generation (v1 06/15)}
% Generative recommendation models adopts sequential generative models to generate personalized recommendations for users based on their preferences, past interactions, and other relevant data. Unlike traditional recommendation systems, generative recommendation systems are able to create new recommendations by modeling the underlying distribution of user-item interactions.
%HSTU is one of the generative recommendation models that reformulates recommendation problems (retrieval, ranking) as sequential transduction tasks within a generative modeling framework. HSTU has demonstrated an impressive model quality improvement for production-scale tasks. It comes with superior scaling performance compared to traditional DLRMs~\cite{dlrm, dlrm2}. These generative recommendation models enables a unified feature space that can be used across various product domains.

%HSTU is the Hierarchical Sequence Transduction Unit (HSTU) model architecture is a type of transformer-based model designed primarily for sequence-to-sequence tasks, such as machine translation or speech recognition.
% HSTU is composed of a stack of identical layers connected by residual connections~\cite{residual}. Each layer consists of three main sub-layers: \textbf{Point-wise Projection}, \textbf{Spatial Aggregation}, and \textbf{Pointwise Transformation}.

% \vspace{-0.28155cm}
% \hstu
% Jiaqi's initial rewritten version below (06/16) (if needed please add jz392@cornell.edu)
\subsubsection{gDLRM for Generative Recommendations}\label{hstuarchitecture}
Generative recommenders approach information retrieval and recommendation problems by modeling the underlying joint distribution of user-item interactions, and adopting homogeneous, large-scale sequential backbones to replace the traditional heterogeneous modules in DLRMs. gDLRMs enables the main tasks in recommendations, namely retrieval (predict the next item to recommend) and ranking (predict the engagement type given the retrieved item), to be formulated as a next-token prediction problem. We refer to both type of outputs for retrieval and ranking tasks as "\textit{Action}".
Compared to prior DLRMs~\cite{dlrm, gupta-hpca2020, dlrm-2, hsia-asplos23}, gDLRMs have demonstrated superior accuracy performance~\cite{hstu} and further enable a unified feature space to be used across different domains.

One key sequential architecture used in the generative recommender system --- HSTU --- can be viewed as a variant of self-attention or Transformers specialized for sequence-to-sequence (sequential transduction) tasks. HSTU is composed of a stack of identical layers connected by residual connections~\cite{residual} as shown in Figure~\ref{fig:hstuarch}. Each layer consists of three main sub-layers: \textit{Point-wise Projection}, \textit{Spatial Aggregation}, and \textit{Pointwise Transformation}. 
%At a high level, 
\textit{Spatial Aggregation} replaces sequence-level normalized Softmax with pointwise normalized attention and relative attention bias, and \textit{Point-wise Projection} together with \textit{Pointwise Transformation} together performs efficient token-level transformation augmented by element-wise gating. 
This reduces the number of matrix multiplication operations from standard Transformers. In general, training throughput performance can be significantly improvement through feature deduplication optimization~\cite{zhao2023recddeduplicationendtoenddeep}. Note that HSTU is the only model that is non-autoregressive among the generation tasks studied in this paper.

\vspace{-0.04in}
\section{System Performance Characterization on Multi-Modal Model Inference}

We present real-system performance characterization results for the key generative AI tasks in this section. The deeper understanding of application-level characteristics and performance bottlenecks on real systems help guide our performance and efficiency optimization focus systematically. The data-driven analysis also underpins key system design and optimization opportunities, as what we later show in Section~\ref{sec:opportunity}.

\begin{figure}
  \centering
     \includegraphics[width=0.6\columnwidth]{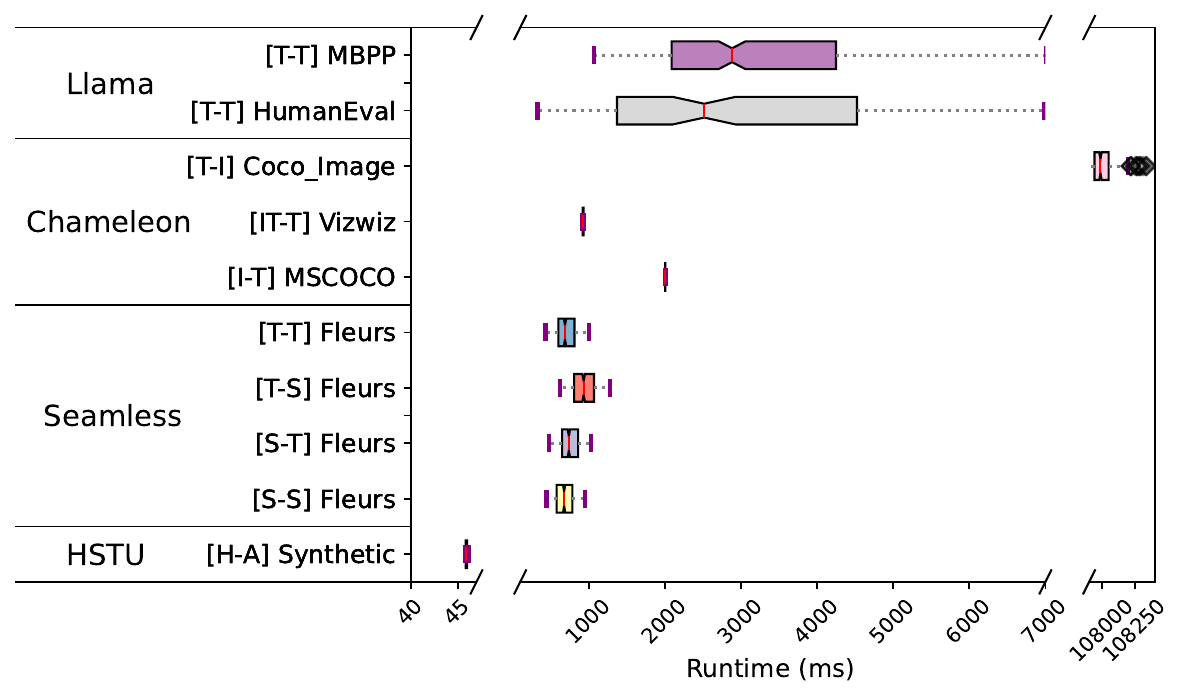}
     % \vspace{-0.5cm}
     \caption{Latency Distribution of each workload.}
     \label{fig:latency_dist}
     \vspace{-0.25cm}
\end{figure}

\subsection{Sequence Length and Latency Distribution}\label{sec:seqlen}
Sequence length is a key task-specific dimension that determines \textit{where} the most important performance acceleration opportunities come from.
Sequence length distribution also affects the computational efficiency of generative models. For instance, models with shorter sequence lengths require less computation time to generate samples compared to models with longer sequence lengths. 
Transformer-based models, in particular, are highly sensitive to sequence length distributions due to their attention operation, where computational costs increase quadratically (i.e., $O(N^2d)$, where $N$ and $d$ denote sequence length and embedding dimension, respectively).
% Understanding the sequence length distribution can help optimize the trade-off between sample quality and computational efficiency.
In Table~\ref{tbl:seq_len}, we delve into the sequence length distribution for the four different generative AI models. 

And in Figure~\ref{fig:latency_dist}, we show end-to-end inference latency distribution to show the correlation between the sequence length and the latency. We measure the inference latency of each sample with a batch size of 1 on NVIDIA A100 GPU to get the latency distribution. Based on our analysis, latency distribution is highly correlated to the sequence length distribution. We are going to discuss the correlation in detail in the following parts. 
By understanding the sequence length and latency distribution and its correlation, our goal is to better understand what determines the different system performance of four generative AI models and help to optimize those models by understanding the trade-off between the length of the sample and the computational efficiency.
% \color{black}

\noindent{\textbf{Llama:} For the Llama-based coding tasks (Code Llama), we focus on coding capabilities of AI using the HumanEval~\cite{data:humaneval} and MBPP~\cite{data:mbpp}, respectively.
%take the first 100 samples of two representative datasets, HumanEval~\cite{data:humaneval} and MBPP~\cite{data:mbpp}, that includes programming problems and are used for evaluating coding capability of AI models. 
The input prompts describe the programming problems in text, such as \textit{"Write a python function to find the first repeated character in a given string."}. 
We define input sequence lengths of Coda Llama as the number of text tokens fed into the model whereas output sequence lengths represent the number of text tokens generated by the model. 
In general, the input sequence length for MBPP is in the order of tens of tokens while the input sequence length for HumanEval is in the hundreds. This is because the HumanEval dataset gives more detailed constraints of problems with simple examples in the input prompts. In contrast, the output sequence lengths for HumanEval is in the order of hundreds since the solution for these datasets are quite simple that could be solvable in few lines of code (around 10 lines in general).}

In Figure~\ref{fig:latency_dist}, we report the latency distribution of HumanEval and MBPP dataset. 
Overall, MBPP has longer end-to-end latency than HumanEval as the number of decoding step is the key factor in deciding the end-to-end latency which will be discussed in more detail in Section~\ref{sec:opportunity} Observation \#1. T-T tasks have the widest latency distribution among all tasks as the end-to-end latency has high correlation with the sequence lengths and the number of decoding steps distribution. The standard deviation is one of the most representative metric to show how broadly the values are distributed. Our observation says T-T tasks have the largest standard deviation for input sequence lengths and decoding steps. %\yejin{Should we add std value to the table?}

% \seamless
\noindent{\textbf{Seamless:} 
For sequence length analysis of Seamless, we focus on the Fleurs~\cite{data:fleurs} dataset which contains the speech version of the FLoRes~\cite{flores} machine translation benchmark in 102 different languages. This dataset is used for a variety of speech tasks, including automatic speech recognition (ASR), speech language identification, translation and retrieval.

The input sequence for the Seamless M4T model is generated by extracting 80-dimensional filterbank features from the raw audio waveform at the 100Hz frame rate and by stacking every 2 frames for the final 160-dimensional 50Hz features. These extracted features, i.e., a dimension of 160 for Seamless M4T, become the input and the number of features becomes the sequence length of the model. 
The input sequence length statistics are for speech encoder in case of S-T and S-S tasks and text encoder in case of T-T and T-S tasks. For output sequence lengths statistics, we report the output sequence lengths of text decoder module in case of text generation tasks (S-T, T-T) and the output sequence lengths of NAR T2U module in case of speech generation tasks (S-S, T-S).

The output sequence of Seamless is specific to the corresponding tasks. For text generation tasks (T-T, T-S), we define the output sequences as the generated text tokens from T2TT module. For speech generation tasks (S-T, S-S), we define the output sequences as generated  units from NAR T2U module.
% Since T2TT is the only autoregressive module in Seamless, we report the input sequence lengths and the number of decoding steps for T2TT module in Table~\ref{tbl:seq_len}. Note that the input sequence length for T2TT is the same for speech encoder module and the output sequence length of T2TT is the same for NAR T2U and Vocoder modules. 
Furthermore,
we take English to Spanish translation as our analysis use case since it is one of the most frequently used combinations for translation task. We use \texttt{en\_us} and \texttt{es\_419} subset from Fleurs dataset for English source language and Spanish target language, respectively.
The average duration of input speech files of \texttt{en\_us} are around 9.88 sec, resulting in the average input sequence length for speech modality as 986 and 30 for text modality.
% In case of text generation tasks, the average output sequence length is XX while speech generation tasks have the average output sequence length XX which generates average XX sec of audio files. 
% The training data for SeamlessM4T is limited to 20 seconds (i.e. 1000 frames).}

In Seamless, text generation tasks only utilize conformer speech encoder and T2TT modules while speech generation tasks run NAR T2U and vocoder in addition. Thus, speech generation tasks generally take longer than text generation tasks. In our analysis, S-S tasks are 24\% slower than S-T tasks and T-S tasks are 20\% slower than T-T tasks on average in terms of inference latency.

% \chameleon
\noindent{\textbf{Chameleon:} For Chameleon-based multi-modal tasks, we focus on the widely-used MSCOCO~\cite{data:mscoco}, Vizwiz~\cite{data:vizwiz} datasets for I-T\&T-I and IT-T tasks, respectively.
\vspace{-0.05in}
\begin{itemize}[leftmargin=*]
    \item \textbf{Image to Text (I-T) tasks:} Chameleon uses newly trained image tokenizers~\cite{cm3imagetokenizer} and the BPE tokenizer~\cite{cm3texttokenizer} for the image and text input modality, respectively.
    The I-T task uses 1030 tokens, combining 1024 image tokens with 6 prompt tokens (e.g., "Describe the figure") for caption generation.
    % The image tokenizer generates 1024 image tokens per image. Thus, the I-T task has a fixed input sequence length of 1030 tokens, including the 1024 tokens representing an image and an additional 6 tokens representing the static prompting telling the model to generate the caption (e.g., "Describe the figure").
    
    \item \textbf{Image/Text to Text (IT-T) tasks:} For the IT-T generation task, a representative use case is Visual Question Answering (VQA), which generates response given an image and a question for the image, such as "Can you tell me what this image is about?". 
    In this case, the input sequence combines 1024 image tokens with additional tokens for the question text. Taking the Vizwiz dataset as an example, the number of text tokens for the questions range from 3 to 65.

    % Note that I-T and IT-T tasks have fixed number of decoding steps because Chameleon decodes until a maximum output length for a given task and then extracts the substring containing the desired predictions using task-specific templates. This is why the number of decoding steps remains fixed, even though the output lengths for different tasks may vary.

    Note that I-T and IT-T tasks maintain fixed decoding steps by using maximum output lengths and task-specific templates for prediction extraction.

    In Figure~\ref{fig:latency_dist}, 
    % I-T tasks generally have longer latency than IT-T tasks even though they have similar input sequence length. This is because the number of decoding steps of I-T tasks are longer than IT-T tasks. The maximum output length of I-T (30) is longer than that of IT-T (10) because generated text caption for image (image captioning) is generally longer than the generated text answers for the image (visual question answering, VQA) because VQA asks questions that can be answered in few words such as "Q: Which one is the blue one? A: Right", "Q: "What color is this A: White".
    I-T tasks have longer output length (30) than IT-T (10) since image captioning require more words than VQA (visual question answering) answers, which are typically brief responses like "Q: Which one is the blue one? A: Right", "Q: "What color is this A: White".

    \item \textbf{Text to Image (T-I) tasks:} For the T-I generation task, instructions to generate image such as "An upstairs living room is decorated nicely and holds a sewing machine." is given as the text input prompt. Thus, the input sequence length is determined by the number of text tokens generated by the text tokenizer. We use the MSCOCO~\cite{data:mscoco} dataset, for which the average input sequence length is 13.9.

    In Figure~\ref{fig:latency_dist}, T-I tasks have the longest latency among the all tasks. Even though T-I tasks have shorter input sequence lengths, the number of decoding is highest which is 1024 resulting in the longest latencies. Also, as mentioned in Section~\ref{sec:chameleonbackground}, Chameleon uses a contrastive decoding methiod for T-I task, thus Chameleon runs the model twice at each incremental decoding step.
\end{itemize}
}

% \hstu
\noindent\textbf{gDLRM:} For recommendation tasks with feature generation (HSTU), we focus on a synthetically generated dataset, where a sequence of user history is randomly generated. We generated 16384 number of inference samples, and the sequence of each sample comes with random integer indices, which range from 0 to 6000. The synthetically-generated sequence lengths are configured to represent the distribution we observed in the production environment as mentioned in the work~\cite{hstu}. % 2048-8192

% Even though the average input sequence length is 3000, the average of generated user history is 4813.9, because 
% \yejin{@Jiaqi Could you fill in the explanation here?}.
Also, as mentioned in Section~\ref{hstuarchitecture}, HSTU is composed of a stack of identical 14 layers but they limit the maximum input sequence length for later 11 layers as 1024 for speed improvement performance.
% \color{black}

% Overall, understanding the sequence length distribution is crucial for optimizing the performance and behavior of generative AI models. By analyzing this property, researchers and developers can gain insights into the strengths and weaknesses of different models and make informed decisions about which models are best suited for specific applications.

% python onellm_scripts/breakdown_script/parse_json_parallel_barebones_enhanced.py --figure1
% \begin{figure*}
% \vspace{0.25cm}
%   \centering
%   \includegraphics[width=\textwidth]{Submission/analysis_figures_revision/breakdown/overall_breakdown_normalized.pdf}
%   \caption{Operator Time Breakdown of Code Llama~\cite{codellama}, Seamless~\cite{seamless}, Chameleon~\cite{chameleon}, HSTU~\cite{hstu}.}
%   \label{fig:genai-model-breakdown}
% \end{figure*}

% \begin{figure*}
% \vspace{0.25cm}
%   \centering
%      \begin{subfigure}[b]{0.49\textwidth}
%          \centering
%          \includegraphics[width=\textwidth]{Submission/analysis_figures_revision/breakdown/seperate_overall_breakdown.pdf}
%          \caption{Unnormalized}
%      \end{subfigure}
%      \hfill
%      \begin{subfigure}[b]{0.49\textwidth}
%          \centering
%          \includegraphics[width=\textwidth]{Submission/analysis_figures_revision/breakdown/seperate_overall_breakdown_ratio.pdf}
%          \caption{Each bar is normalized}
%      \end{subfigure}
%      \caption{Operator Time Breakdown - Prefill and Decode stages are seperated for Chameleon and CodeLlama}
% \end{figure*}

\begin{figure*}
\vspace{0.25cm}
  \centering
  \includegraphics[width=0.8\textwidth]{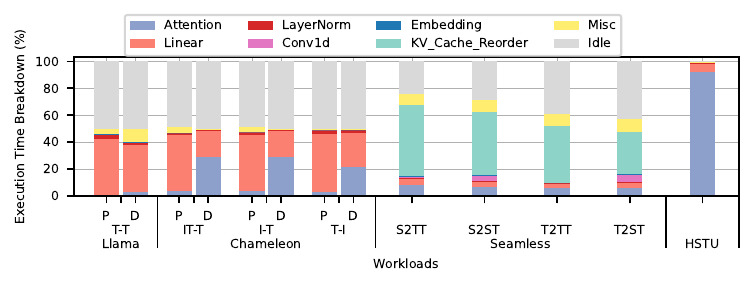}
  % \vspace{-0.25in}
  \caption{Operator Time Breakdown of Llama~\cite{codellama}, Seamless~\cite{seamless}, Chameleon~\cite{chameleon}, HSTU~\cite{hstu}. \textbf{P} stands for the Prefill stage whereas \textbf{D} stands for the Decoding stage, respectively.}
  \label{fig:genai-model-breakdown}
  \vspace{-0.25cm}
\end{figure*}

Based on the fact that sequence length distribution is unique to each generative AI task, in the next section, we delve into understanding \textit{where} inference latency comes from.
% \yejin{Some of these generative AI models is small in size, so realistically speaking, if we care about resource efficiency, we would try to maximize the batch size to inference latency target and also we try to maximize gpu utilization. When we start doing that, this~\ref{fig:radar_max_batch} is what we get. What is the maximum batch size that can be supported given the gpu capacity that we have. This is  }

\subsection{Operator Time Breakdown}\label{sec:opportunity}
System performance bottlenecks are distinct across the key generative AI tasks Llama (CodeLlama), Chameleon (CM3), Seamless (Seamless), and generative DLRM (HSTU). Depending on the input and output modality types and the corresponding sequence lengths, system performance optimization opportunities vary.

Figure~\ref{fig:genai-model-breakdown} presents the end-to-end model inference time breakdown for Llama, Seamless, Chameleon and gDLRM. 
% By performing in-depth examination of system performance, we aim to better understand different system implication of four different generative models
We characterize the inference time by maximizing the batch size for each workload to fit in the HBM memory capacity (i.e., 80GB) of a single NVIDIA A100 GPU~\cite{nvidiaa100}.
We report the averaged breakdown result for 5 samples after 15 iterations of warmup for Code Llama, Seamless, Chameleon and HSTU (3 samples for T-I task of Chameleon model).
For detailed description of the codebase and environment setup, please refer to Section~\ref{sec:result}.

Four generative AI models consist of different sets of operators. "Idle" time indicates idle time on GPU device during the inference when GPU is in the idle status because of the GPU kernel launch overhead on CPU side. We divide prefill and decoding stage for Llama and Chameleon to better understand the different characteristics of each stage. And we show the normalized inference time to the end-to-end prefill time of Llama on top of each bar. Note that we exclude embedding table lookup time of HSTU given that DLRM serving disaggregates embedding from the main model itself.

\textbf{Observation \#1} \ul{The auto-regressive nature of token generation in Llama and CM3 makes token generation (decoding) a performance-critical phase that is primarily determined by the number of decoding steps, whereas the inference latency of HSTU is much faster and does not depend on token generation.}

For autoregressive generative models (Llama, Seamless, and Chameleon), the number of decoding steps matters the most to the end-to-end latency. As these models generate tokens sequentially, larger the number of decoding steps prolongs the generation process. For example, I-T task and IT-T task have similar average input sequence length while I-T task have 3 times higher number of decoding step according to Table~\ref{tbl:seq_len}. This result in longer end-to-end inference latency as shown in Figure~\ref{fig:genai-model-breakdown}. Also, the T-I task in the Chameleon model takes the longest latency per inference sample because the image generation process involves 1024 decoding steps to produce a single image, significantly larger than the number of steps required by other tasks. This results in the longest latency per inference sample among the four models. 
Also, Llama has longer latency compared to  I-T task and IT-T task of Chameleon even the input sequence lengths for Llama is much smaller (upto 13x). One of the primary cause for this is because Llama has higher number of decoding steps, resulting in the increased end-to-end latency.

% For autoregressive models, the prefill stage takes the full input prompt and an incremental decoding stage takes the previously generated output and generates output tokens one by one. In incremental decoding stage, KV cache optimization is adopted to relieve computational overhead making computation to happen only for a single token that is generated in the previous incremental decoding step. Thus the input sequence length for the incremental decoding stage is always 1 with KV cache optimization while the input sequence length for the prefill stage is determined by the length of the full input prompt.
Considering that the prefill stage is only performed once while the incremental decoding stage is repeated multiple times, the number of decoding steps has a more significant impact on end-to-end inference latency than the input sequence length of the prefill stage when the number of decoding steps is non-trivial.

On the other hand, non-autoregressive models generate all tokens simultaneously rather than sequentially, so they can be significantly faster than autoregressive models. This is particularly beneficial for long sequences or when real-time performance is crucial and this can lead to a better user experience. HSTU~\cite{hstu} demonstrates the potential benefits of non-autoregressive models.

\textbf{Observation \#2} \ul{The inference time of autoregressive models is often dominated by the GPU idle time, indicating that these models depend heavily on CPU-bound modules.}

We observed a significant gap incurred by CPU overhead that delayed the launch of GPU kernels, resulting in GPU underutilization and a substantial increase in the execution time especially for Llama and Chameleon. 

Seamless and HSTU have relatively higher GPU utilization compared to Llama and Chameleon. For Seamless, Speech/Text Encoder and Text Decoder are always activated and NAR T2U and Vocoder are selectively activated depending on the tasks. Among the four modules, only the text decoder is an autoregressive module indicating that only this module will be operating on matrix size of sequence length 1 except for Prefill while the rest modules are operating on the matrix size of the given full input sequence length. Thus, the overall GPU utilization for Seamless is higher than Llama and Chameleon since it has only one autoregressive module.
For HSTU, the input sequence length is much larger (4813.9 x batch size) than other models according to Table~\ref{tbl:seq_len}, so GPU spends much more time on computation resulting in high GPU utilization.

To address CPU-bound issue, optimizing techniques like torch.compile and CUDA Graph can significantly reduce the GPU kernel launch overhead. The latency improvement results of torch.compile and CUDA Graph are provided in Section~\ref{sec:torch-compile}.

\textbf{Observation \#3} \ul{Across all workloads, linear operations constitute a comparable portion of the overall model inference latency as the attention operations due to the Feed Forward Networks (FFNs) in Transformer-based models.}

For Llama and Chameleon, Linear operation dominates the end-to-end inference time.
For Seamless, linear operation takes a comparable portion of the inference time to attention operation. And for HSTU, attention operation dominates the inference time, unlike the other models. That is because the computation cost of attention operation grows quadratically ($O(N^2)$) to the input sequence length and the input sequence length of HSTU is in higher order than other generative models as addressed in Table~\ref{tbl:seq_len}. 

Generally, the linear operation takes an insignificant amount of the inference time, thus accelerating linear layer operations could bring significant improvements to end-to-end latency than accelerating attention operations. In Section~\ref{sec:autoquant}, we delve more deeply into inference acceleration using different numeric precision levels on the linear operation performance and output quality.

% could lead to better trade-offs between speed and accuracy. We report the latency improvement results and detailed explanations of the quantization technique in Section~\ref{sec:autoquant}. 

\textbf{Observation \#4} \ul{KV Cache reordering operation dominates Seamless inference time, which is a necessary operation for the decoding strategy based on beam search.} 

Autoregressive models perform incremental decoding steps based on the decoding strategy that the model adopts. Decoding strategy is a sampling method used to choose the next token based on the output probability distribution over the vocabulary dictionary.
The popular decoding strategies include deterministic methods such as greedy and beam search and stochastic (sampling) methods such as top-p, top-k, random, etc. Llama and Chameleon use top-p decoding strategy and Seamless uses beam search decoding strategy. Beam search decoding strategy is widely used for closed form generation tasks such as translation, because sampling based decoding strategies are way too stochastic which often lead to a worse semantic match between the predicted and reference sequence.

Beam search maintains a beam of the \texttt{K} best sequences so far and considers the probabilities of the combination of all of the preceding words along with the word in the current position. Beam search maintains a separate copy of the KV cache for each sequence, and it needs to reorder KV caches for all attention layers according to the selected sequences from the previous decoding step to make sure each selected beam performs with the corresponding KV cache. This step requires copying all KV cache into a new memory space resulting in a significant portion of the inference runtime. This could be further optimized with torch.compile and we discuss the torch.compile case study for Seamless in Section~\ref{sec:torch-compile}.

\begin{table}[]
\centering
\small
\begin{tabular}{l|l|l|l|l}
\makecell[l]{Model \& Codebase}          & Task                      & Dataset                 & \makecell[l]{Max.\\Batch Size} & \makecell[l]{\# of\\Samples} \\\hline\hline
Llama~\cite{bigcode-evaluation-harness} & T-T                       & HumanEval               & 4                              & 164                          \\ \hline
\multirow{3}{*}{\makecell[l]{Chameleon~\cite{chameleon}}}                 & I-T                       & MSCOCO                  & 16                             & 5000                         \\ \cline{3-5}
                                                               & IT-T                      & Vizwiz                  & 16                             & 4319                         \\ \cline{3-5}
                                                               & T-I                       & Coco Img.              & 16                             & 500                          \\ \hline
\multirow{2}{*}{\makecell[l]{Seamless~\cite{seamlessgithub}}} & S-S \& S-T             & \multirow{2}{*}{Fleurs} & 128                            & \multirow{2}{*}{643}         \\ \cline{4-4}
                            % &                                      & S-T                    &                         & 128                            &                              \\ \cline{5-5}
                                                                  & T-T \& T-S             &                         & 384                            &                              \\ \hline
                            % &                                      & T-S                    &                         & 384                            &                              \\ \hline
HSTU~\cite{hstu}                                    & H-A                    & Synthetic               & 32                             & 16384               \\ \hline
 
\end{tabular}
\caption{Datasets, codebase and batch size configuration.}
\label{tbl:tasks}
% \vspace{-0.4in}
\end{table}

% \begin{table}[]
% \centering
% \small
% \begin{tabular}{l|l|c}
% Category                                                  & Optimization           & Impact to Accuracy \\\hline\hline
% \multirow{3}{*}{\makecell[l]{System-level Optimization}}  & SDPA                   & \tikzxmark   \\ \cline{2-3}
%                                                           & torch.compile          & \tikzxmark   \\ \cline{2-3}
%                                                           & AutoQuant              & \tikzcmark   \\ \hline
% \multirow{2}{*}{Workload-specific Optimization}           & LayerSkip              & \tikzcmark   \\ \cline{2-3}
%                                                           & CHAI                   & \tikzcmark   \\ \hline
% \end{tabular}
% \caption{Accuracy Impact of Optimization Techniques.}
% \label{tbl:accuracyimpact}
% \vspace{-0.5cm}
% \end{table}

\vspace{-0.1528in}
\section{Accelerating Multi-Modal Model Inference via Cross-Stack Optimization}\label{sec:result}

In this section, we highlight the importance of enhancing inference performance by taking into account state-of-the-art system optimization techniques as well as algorithmic advancement. 
% The adoption of these advanced optimization techniques not only accelerates technological advancement but also fosters a collaborative research environment conducive to cumulative progress. By embracing such established methodologies, researchers establish a robust foundation for enhancing system efficiency, mitigating risks associated with inefficiencies and unforeseen bottlenecks. 
There are (1) horizontal system-level optimizations and (2) vertical workload-specific optimization techniques. 
\begin{itemize}[leftmargin=*]
    \item System-level techniques optimize inference time performance horizontally across the generative AI tasks while being agnostic to specific algorithms. We consider Scaled Dot Product Attention (SDPA), torch.compile and CUDA graph optimization (CUDA Graph). 
    %in Section~\ref{sec:torch-compile}. 
    SDPA leverages highly optimized and fused implementation to reduce the number of kernel launches and intermediate data transfers, which contributes to lower latency and memory usage.
    torch.compile and CUDA Graph facilitate streamline GPU task scheduling and execution, optimizing parallelism and resource utilization given system hardware. 
    We deploy quantization optimization using the PyTorch AutoQuant framework~\cite{torchao} --- in Section~\ref{sec:autoquant}. It automates the tuning process by determining the most efficient quantization method for each layer. 
    % by itself and ensures optimal system performance across diverse workloads, circumventing the need for manual intervention.

    \item Workload-specific techniques optimize design objectives tailoring to algorithm or neural network (NN) specific characteristics. Taking a recent NN optimization technique ---  LayerSkip~\cite{layerskip}
    % and CHAI~\cite{chai}
    --- tailor-designed for Transformer-based large language models, we evaluate the impact of LayerSkip across the generative AI tasks in Section~\ref{sec:worklaodspecific}
    % in this paper. 
    % LayerSkip adopts the idea of skipping layers in the model to reduce inference time.
    % while CHAI exploits similarity patterns by clustering attention heads in the multi-head attention operation to speed up inference. Section~\ref{sec:worklaodspecific} presents the inference time improvement results across the family of the key multi-modal generation tasks. 
\end{itemize}

\noindent\textbf{Metholody Detail:} Table~\ref{tbl:tasks} presents the datasets and the corresponding codebase used for each task, as well as the maximum batch size that fits in a single NVIDIA A100 GPU~\cite{nvidiaa100} used in our study. For the MSCOCO image dataset, we sub-sampled 500 out of 2000 data samples so the experiment time is more manageable and used the full dataset for the rest tasks. For HSTU, we generated synthetically a dataset with 16,384 samples, where a sequence of user history for each sample is randomly generated as explained in Section~\ref{sec:seqlen}. We validated and ensured the dataset is representative of production usecases.

\vspace{-0.1in}
% \xformer
\subsection{Baseline is All You Need}\label{sec:baselineisallyouneed}

\subsubsection{Accelerating Attention}\label{sec:sdpa}
The Attention operation in Transformer-based model architectures is an Amdahl's law bottleneck. Based on the real-system performance characterization in Figure~\ref{fig:genai-model-breakdown}, Attention contributes to 3.4\% of the end-to-end inference time in the decoding phase for Code Llama whereas, for HSTU, over 90\% of the inference time comes from the Attention operation. 

To accelerate Attention, we enable PyTorch SDPA (Scaled Dot Product Attention)~\cite{sdpa} designed specifically to accelerate the fundamental building block --- Attention --- in Transformer-based model architectures. PyTorch provides  \texttt{\footnotesize torch.nn.functional.scaled\_dot\_product\_attention} as a function to optimize the inference time performance by accelerating the dot product computation between the Query, Key, and Value matrices using SDPA~\cite{attentionisallyouneed}. 
% While this function can be written in PyTorch using existing functions, PyTorch SDPA function provides a fused implementation which can provide large performance benefits over a naive implementation.

% Depending on tensor inputs, the function dispatches one of the following implementations:  FlashAttention-2~\cite{flashattention2}, Memory-Efficient Attention~\cite{xformers}, or a PyTorch implementation defined in C++. The function may call optimized kernels for improved performance when using the CUDA backend. For all other backends, the PyTorch implementation will be used. All implementations are enabled by default. Scaled dot product attention attempts to automatically select the most optimal implementation based on the inputs. Each of the fused kernels has specific input limitations.

Instead of relying on the PyTorch SDPA API directly, for HSTU, we manually implemented the memory-efficient attention~\cite{rabe2022selfattentiondoesneedon2} and Flash Attention~\cite{flashattention} as is in PyTorch, directly at HSTU's internal code base. The memory-efficient attention implementation divides the input into blocks and avoid materializing the large $h \times N \times N$ intermediate attention tensors for the backward pass. This reduces the attention computation as a group of back-to-back GEMMs \textit{with different shapes}, which enables the sparsity of input sequences to be exploited. 
% achieving upto 15x speedups on 8K sequences. 
The construction of the relative attention bias is also a bottleneck due to memory accesses. To address this issue, we fused the relative bias construction and grouped GEMMs into a single GPU kernel, and accumulates gradients using GPU's fast shared memory in the backward pass.

\begin{figure}
     \centering
    \includegraphics[width=0.65\columnwidth]{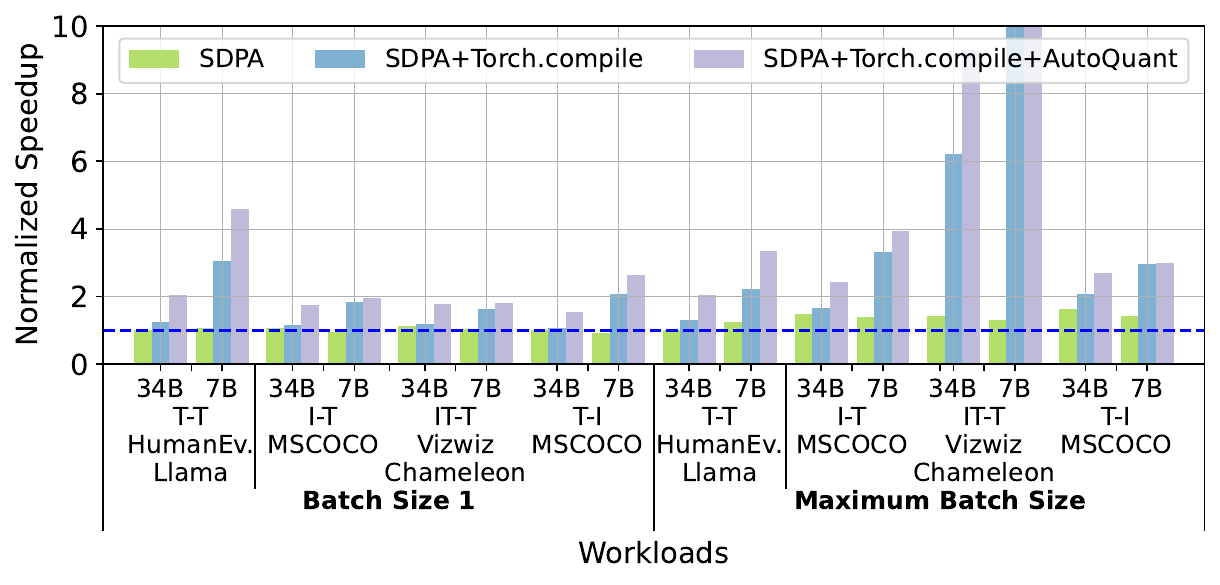}
     \caption{End-to-end inference time speedup with SDPA and torch.compile for Llama and Chameleon on A100 GPU.}
     \label{fig:sdpa}
     \vspace{-0.1in}
\end{figure}
\noindent \textbf{Results -- SPDA.} Figure~\ref{fig:sdpa} presents the end-to-end latency speedup across the family of multi-modal generation tasks for the settings of batch size=1 and of the maximum batch size (the largest batch size that can support each model on a single A100 NVIDIA GPU as configured in Table~\ref{tbl:tasks}). 
PyTorch SDPA accelerates inference time performance of the generation tasks by an average of \sdpabsone and \sdpabsmax for the single-batch and maximum-batch settings, respectively.
In particular for HSTU, using the same fundamental principle, we observe 2.11$\times$ and 9.87$\times$ inference time improvement for the single-batch and maximum-batch settings, respectively. The significant inference time speedup stems from the proportionally-larger amount of time spent on the Attention operation for HSTU than the other generation tasks. And we observed that HSTU optimized implementation achieves up to 15$\times$ speedups on 8K sequences.

% In general, PyTorch SDPA sets a more competitive baseline for inference performance across the board. However, we observe that there could be no performance gain for the cases where attention operation accounts for significantly lower than other tasks. For example, we observe no performance gain for using SPDA for Seamless because Seamless has the smallest portion of attention runtime among four generative AI models,  which is less than 7\% across all tasks.

In general, PyTorch SDPA generally establishes a more competitive baseline for inference performance across all tested scenarios. However, it's important to note that performance gains may be negligible in cases where the attention operation constitutes a significantly smaller proportion of the overall inference runtime. For instance, we observed no performance improvement when applying SDPA to Seamless, as it allocates the smallest portion of runtime to attention operations among the four generative AI models examined—less than 7\% across all tasks accordign to Figure~\ref{fig:genai-model-breakdown}.

\begin{figure}
     \centering
     \begin{subfigure}[b]{0.42\columnwidth}
         \centering
         \includegraphics[width=\textwidth]{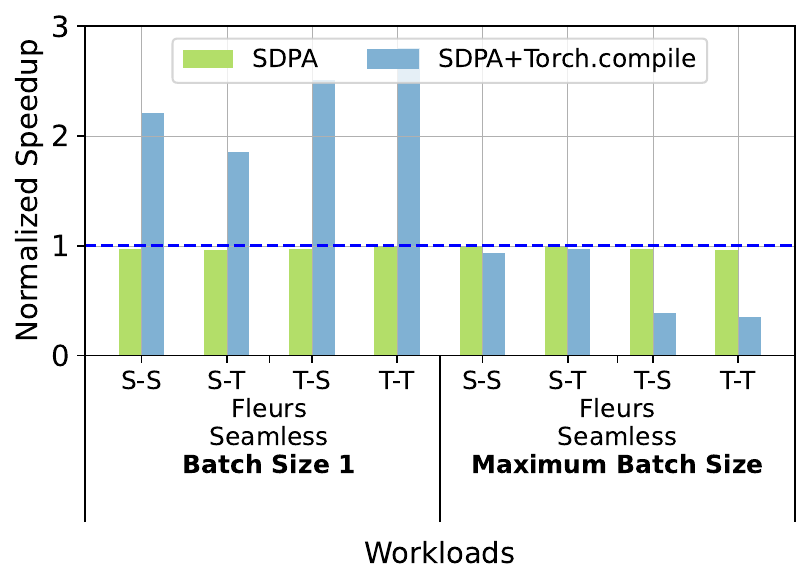}
         \caption{Seamless}
     \end{subfigure}
     % \hfill
     \begin{subfigure}[b]{0.22\columnwidth}
         \centering
         \includegraphics[width=\textwidth]{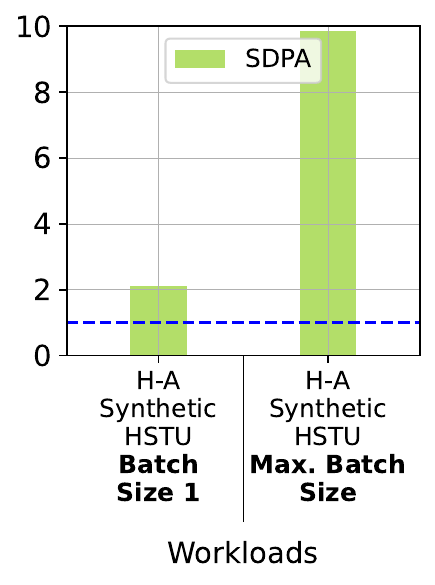}
         \caption{HSTU}
     \end{subfigure}
     \caption{End-to-end inference time speedup with SDPA and torch.compile for Seamless and HSTU on NVIDIA A100 GPU.
     %excluding AutoQuant optimization as linear operation is not a bottleneck according to Figure~\ref{fig:genai-model-breakdown}. Refer to Section~\ref{sec:autoquant} for more details.
     }
     \label{fig:torchcompilespeedup}
     \vspace{-0.25in}
\end{figure}

% \begin{figure*}
%      \centering
%      \begin{subfigure}[b]{0.49\textwidth}
%          \centering
%          \includegraphics[width=\textwidth]{Submission/analysis_figures_revision/torch_compile/batch_size_1.pdf}
%          \caption{Batch Size 1}
%      \end{subfigure}
%      \hfill
%      \begin{subfigure}[b]{0.49\textwidth}
%          \centering
%          \includegraphics[width=\textwidth]{Submission/analysis_figures_revision/torch_compile/batch_size_max.pdf}
%          \caption{Maximum Batch Size}
%      \end{subfigure}
%      \caption{End-to-end inference time speedup with torch.compile and AutoQuant.}
%      \label{fig:torchcompile2}
% \end{figure*}

\subsubsection{Improving GPU Utilization}\label{sec:torch-compile}
During inference, for the single batch setting, the workloads are typically not compute bound, which raises 2 issues. First, each kernel that runs on the GPU becomes so fast, the overhead of launching kernels starts dominating the overall inference time. We reduce the number of kernels with PyTorch's compiler. \texttt{torch.compile}~\cite{pytorch-2.0} accelerates PyTorch models by capturing and optimizing their computation graph. This includes fusing multiple operations into a single kernel.
The second and more important issue of inference is that the GPU computations can be faster than the time it takes to execute the corresponding python code on CPU. The consequence is that the GPU is inactive most of the time, waiting for instructions from the CPU. We address this with CUDA graphs~\cite{cudagraph}.
A CUDA graph is a succession of GPU operations that can be executed as a whole, without having to execute CPU code to schedule kernels one-by-one. In particular, this ensures that the GPU is always active during the graph execution. In practice, the graph is captured once when running the PyTorch model, and can then be replayed whenever we have a new input.
%This can lead to significant performance improvements, especially for applications that have a large number of small kernels or repeat the same set of kernels multiple x. 

% As CUDA Graph is supported as part of torch.compile, it is one of the most widely used optimization techniques for AI models running on GPUs.
One key limitation is that the operations must be in exactly the same static tensor shapes with the same memory addresses. This is incompatible with inference workloads, because the KV cache increases with each iteration, as tokens get appended (\texttt{cache=torch.cat((cache, new\_value), dim=0)}).
To enable CUDA Graph under this limitation, we deployed a static buffer for the KV cache with the maximum sequence length supported by the model prior to the inference. As new keys and values are added to the cache, we increment the current token position on a GPU tensor. This counter is used by the kernel that copies the new tokens inside the KV cache. It is also used by the attention kernel, to skip the part of the KV cache that is not filled yet. This change enables CUDA Graphs, since now the KV cache and the counter have a static shape with a static GPU memory address.
Note the baseline we compare with adopts the optimized implementation with a dynamic KV cache.

\noindent \textbf{Results -- torch.compile/CUDA Graph.} Figure~\ref{fig:torchcompilespeedup} presents the additional inference performance speedup with torch.compile and CUDA Graph.
% Note, the baseline here considers the improvement from PyTorch SDPA in Section~\ref{sec:sdpa}. 
%We evaluate for two configurations a) batch size 1 across all workloads and b) maximum batch size for each workload that fits in a single A100 NVIDIA GPU as configured in Table~\ref{tbl:tasks} and also for two model sizes for Chameleon and Llama which are 7B and 34B to show the different impact of torch.compile depending on the compute intensity.
Overall, the end-to-end inference performance sees an average of \compileoversdpabsone~and \compileoversdpabsmax~additional speedup on top of sdpa for the two batch settings, respectively. This results in total \compileovervanillabsone~and \compileovervanillabsmax~speedup over the baseline without any optimization.

% An exception, we observed the performance degradation for Seamless at maximum batch size due to the requirement of CUDA graph for for static KV cache. As mentioned earlier, substantial speedups are typically expected when static KV cache is used in conjunction with CUDA Graph despite the increased computational demands. However, this is an good example of showing where the computational cost increase associated with static KV cache surpasses the benefits provided by CUDA Graph.
An exception, we observed performance degradation for Seamless at maximum batch size due to CUDA graph's requirement for static KV cache. As mentioned earlier, substantial speedups are typically expected when static KV cache is used with CUDA Graph despite increased computational demands. However, this is an example of where the computational cost increase from static KV cache surpasses the benefits provided by CUDA Graph.

% As shown in Table~\ref{tbl:tasks}, Seamless accommodates the largest maximum batch size on a single GPU among the four generative AI workloads examined. This is due to Seamless having the smallest model size at 2.3B parameters. Our observations indicate that Seamless reaches a performance shifting point as the batch size approaches its maximum, while it is not observed in other models. Furthermore, the advantages of CUDA Graph tend to diminish as models become increasingly compute-bound with larger batch sizes.
% This case study exemplifies the critical importance of understanding workload characteristics when selecting and applying optimization techniques, even within the context of a single model architecture.

\begin{figure}
  \centering
  \includegraphics[width=0.65\columnwidth]{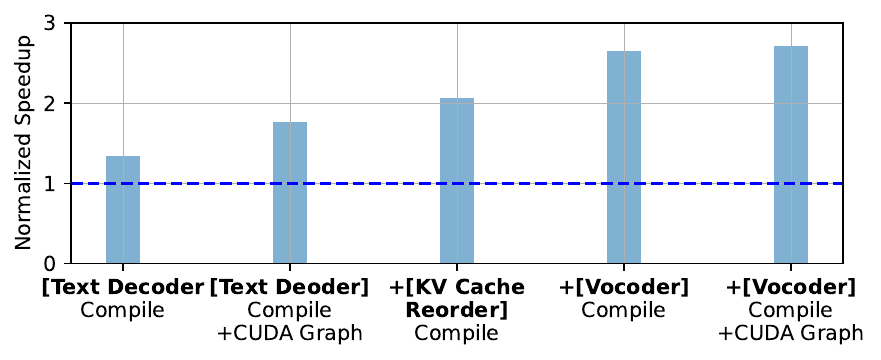}
  \caption{
  End-to-end inference speedup of applying torch.compile incrementally on NVIDIA A100 GPU.
  % \textbf{a)} “Inc. Decoding”: Apply torch.compile only to the text decoder \textbf{b)} “Inc. Decoding w/ CUDA Graph”: Apply torch.compile + CUDA Graph to the text decoder \textbf{c)} “+KV Cache Reordering”: Additionally apply torch.compile to KV cache reordering operation upon b) \textbf{d)} “+Vocoder”: Additionally apply torch.compile to the vocoder upon c) \textbf{e)} “+Vocoder w/ CUDA Graph”: Additionally apply torch.compile + CUDA Graph to the vocoder upon d).
  }
  \label{fig:torchcompile}
  % \vspace{-0.2in}
\end{figure}

\noindent\textbf{\textit{A Deeper Dive with Seamless}:}
Seamless is an emerging speech translation technology that is important to many product surfaces but has not received similar amount of attention as LLMs nor deep learning recommendation models. We focus significant performance acceleration efforts to enable real-time speech translation built on Seamless and present our key findings in a deeper dive. 

% As Figure~\ref{fig:genai-model-architectures}(b) shows , there are four primary modules in Seamless. 
% The T2TT decoder and vocoder are the most time-consuming modules, accounting for 61\% and 23\% of the end-to-end model inference time, respectively. Enabling torch.compile (mode="max-autotune")  and CUDA Graph for the T2TT decoder and vocoder achieves 2$\times$ speedup for the text decoder and 30$\times$ speedup for the vocoder. This leads to 2.65$\times$ faster end-to-end inference latency.
% It turns out, in particular for single batch inference, GPU kernel launch time is hardly amortized, leading to substantial GPU idle time. Enabling torch.compile without CUDA Graph leaves the performance acceleration potential to to 1.17$\times$ and 18.4$\times$ for the text decoder and the vocoder, respectively. While still significant, it shows the important role of CUDA Graph.

There are four primary modules in Seamless (Figure~\ref{fig:genai-model-architectures}(b)).
Enabling torch.compile (mode="max-autotune") and CUDA Graph for the T2TT decoder and vocoder (the most time-consuming modules, 61\% and 23\% of inference time) achieves 2$\times$ speedup for the text decoder and 30$\times$ speedup for the vocoder. This leads to 2.65$\times$ faster end-to-end inference latency.
It turns out, in particular for single batch inference, GPU kernel launch time is hardly amortized, leading to substantial GPU idle time. Enabling torch.compile without CUDA Graph leaves the performance acceleration potential to to 1.17$\times$ and 18.4$\times$ for the text decoder and the vocoder, respectively. While still significant, it shows the important role of CUDA Graph.

% We also report the speedups for text decoder and vocoder using torch.compile without CUDA Graph, which is supported by torch.compile’s API (i.e., torch.compile(mode="max-autotune-no-cudagraphs")), to identify the impact of CUDA Graph on the performance. Without CUDA Graph, the speedup for text decoder and vocoder reduces to 1.17x and 18.4x. While still quite significant, it indicates the important role of CUDA Graph. We conclude that Seamless M4T-v2 is exposed to a lot of time launching CUDA kernels, especially when we use small batch size (e.g., 1) where the GPU kernel execution time is not long enough to amortize the GPU kernel launch time.

% While torch.compile and CUDA Graph are key to inference time improvement, our detailed operator time breakdown in Figure~\ref{fig:genai-model-breakdown} illustrates that Seamless spends significant amount of time on KV cache management (\texttt{KV\_Cache\_Reorder}). This is because Seamless adopts beam search as a text decoding strategy. 
Our operator time breakdown in Figure~\ref{fig:genai-model-breakdown} illustrates that Seamless also spends significant amount of time on KV cache management (\texttt{KV\_Cache\_Reorder}). This is because Seamless adopts beam search as a text decoding strategy. 
In each incremental decoding step, beam search picks the 'N' beams containing the best sequences so far based on the probabilities of the combination of all of the preceding words + current word. 
For each incremental decoding step, KV cache reordering is needed by all Attention layers to ensure that newly selected beams perform on the corresponding KV caches from previous decoding step --- \texttt{kv\_cache = kv\_cache.index\_select(new\_beams)}. 
This code allocates new memory space and overwrites the memory pointer for \texttt{kv\_cache}. To enable torch.compile for \texttt{KV\_Cache\_Reorder}, we had to modify KV cache reordering to keep the memory pointer of each cache as was recorded by using \texttt{torch.Tensor.copy\_} operator. By enabling torch.compile, all GPU kernels related to reordering are fused and compiled, resulting in the final speedup for Seamless.

%Additionally, Seamless also spends significant amount of time for KV cache management (\texttt{KV\_Cache\_Reorder}) as shown in Figure~\ref{fig:genai-model-breakdown}. Seamless adopts beam search as a text decoding strategy and during the beam search process, we need to perform KV cache reordering for all the attention layers for each incremental decoding step to make sure each selected beam performs with corresponding KV cache by performing \texttt{kv\_cache = kv\_cache.index\_select(new\_order)}. This code allocates new memory space and overwrites the original memory pointer for \texttt{kv\_cache}. Thus we applied torch.compile to \texttt{KV\_Cache\_Reorder} by modifying KV cache reordering to keep the memory pointer of each cache as was recorded by using \texttt{torch.Tensor.copy\_} operator.

Figure~\ref{fig:torchcompile} presents the overall inference speedup we achieve for Seamless step-by-step and Table~\ref{tbl:seamlesslabel} describes the each label used for the figure.  While application-specific performance optimization, such as incremental decoding and KV cache reordering, is important, significant inference acceleration potential can be further achieved by torch.compile and CUDA Graph optimization. For Seamless M4T, an end-to-end inference speedup of 2.7$\times$ can be achieved for the challenging single-batch setting. This is key to efficiently enable low-latency, real-time speech translation tasks.

\begin{table}[]
\centering
\small
\begin{tabular}{l|l}
Label           & Description \\\hline\hline
\multirow{2}{*}{\makecell[l]{\textbf{[Text Dec.}\\Compile}}  & \multirow{2}{*}{Apply torch.compile to the text decoder}   \\
& \\\hline
\multirow{2}{*}{\makecell[l]{\textbf{[Text Dec.]} Compile\\+ CUDA Graph}}  & \multirow{2}{*}{\makecell[l]{Apply torch.compile+CUDA Graph to \\the text decoder on top of above row }}   \\
& \\\hline
\multirow{2}{*}{\makecell[l]{+\textbf{[KV Cache}\\\textbf{Reorder]} Compile}}  & \multirow{2}{*}{\makecell[l]{Apply torch.compile to KV cache\\reordering on top of above row}}   \\ & \\\hline
\multirow{2}{*}{\makecell[l]{+\textbf{[Vocoder]}\\Compile}}  & \multirow{2}{*}{\makecell[l]{Apply torch.compile to the vocoder\\on top of the above row}}   \\
& \\\hline
\multirow{2}{*}{\makecell[l]{+\textbf{[Vocoder]} Compile\\+ CUDA Graph}}  & \multirow{2}{*}{\makecell[l]{Apply torch.compile + CUDA Graph\\to the vocoder on top of above row}}   \\
& \\\hline
\end{tabular}
\caption{Table for description of the labels used in Figure~\ref{fig:torchcompile}.}
\label{tbl:seamlesslabel}
\vspace{-0.2in}
\end{table}

% \yejin{For maximum batch size configuration, we experience slowdown.... Probably because of the padding}

% \begin{figure}
%      \centering
%      \includegraphics[width=\columnwidth]{Submission/analysis_figures_revision/hstu.pdf}
%      \caption{End-to-end Inference Time Speedup of Triton Implementation over PyTorch Implementation for HSTU.}
%      \label{fig:hstu}
% \end{figure}

% \color{black}

\subsection{Data Type Optimization}\label{sec:autoquant}
% https://github.com/pytorch/ao/blob/main/torchao/quantization/autoquant.py#L441-L495

% https://github.com/pytorch/ao/tree/main/torchao/quantization

% 
Quantization is an important optimization before models are deployed for downstream inference. To understand the potential of quantization capabilities, we assess data type optimization by applying AutoQuant (Auto-Quantization)\cite{torchao}. 
% \charles{not sure the best way to refer to it, maybe the AutoQuant tool?}
AutoQuant is a recently developed quantization implementation within the PyTorch torchao library~\cite{torchao} designed to integrate high-performance custom data types, layouts, and kernels into PyTorch workflows. AutoQuant optimizes the quantization process by determining the most efficient quantization for each model layer.
% from a set of potential qtensor subclasses \charles{I don't know if you need to talk about qtensor subclasses, it can be said that it just picks the best quantization method for each model layer, the low level mechanics are probably not relevant}. 
It supports two quantization types --- int8 dynamic quantization, int8 weight-only quantization.%, and int4 weight-only quantization. 
% \charles{int4 weight-only quantization is not an option in autoquant, however leaving the layer un-quantized is. This is a practical limitation not a theoretical one, int4 is not included in autoquant because its much less accurate and much faster than the int8 techniques in general. If you can accept that level of accuracy degradation you are probably best using int4 all the way through, its only when you need int8 level accuracy/perf that optimizing by layer becomes attractive.}

% \charles{i might specify the metric being optimized here is runtime/performance, theoretically other techniques might perform better at reducing peak memory/accuracy, autoquant ONLY attempts to optimize runtime}
Depending on downstream tasks, models of different input modalities, architectures and layer specifications can be quantized in distinct ways. For compute-intensive models, dynamic quantization tends to be most effective as it replaces expensive floating-point matrix multiplication operations with faster integer versions. In contrast, weight-only quantization is more beneficial for memory-bound scenarios, where the primary advantage is reduced weight data loading rather than decreased computational demand. % However, note that AutoQuant focuses on optimizing runtime performance rather than reducing memory requirement.

We enable AutoQuant as follows. First, in the model preparation step, linear layers within a model is identified as candidates for quantization. Then, in the shape calibration step, the model with one or more inputs is profiled for the shape and data types of activations recorded for subsequent uses. Finally, the timing performance of the recorded shape and data types are measured, and the fastest quantization setting is applied to speed up model inference.

AutoQuant is designed to work in conjunction with torch.compile, utilizing \texttt{max-autotune} setting to optimize quantization and achieve maximum performance gains. The quantization kernels within AutoQuant rely on torch.compile to generate high-performance kernels, therefore models must first be adapted to use static KV cache and static memory as highlighted in Section~\ref{sec:torch-compile}.

\noindent \textbf{Results -- AutoQuant.} Figure~\ref{fig:torchcompilespeedup} presents the inference time speedup for AutoQuant. AutoQuant provides additional \autoquantovercompilebsone, \autoquantovercompilebsmax~performance improvements for single batch setting on top of torch.compile (Section~\ref{sec:torch-compile}).
Compared to the baseline without any optimization, we observe an average of \autoquantovervanillabsone~and \autoquantovervanillabsmax~latency improvement for the single and the maximum batch settings, respectively. 
%We focus on two model sizes of the Code Llama and Chameleon tasks---7 billion (7B) and 34 billion (34B) parameter model sizes to demonstrate the varying impact of AutoQuant for models with varying compute intensity.
%The models achieve a speedup of 1.53x and 2.28x for the 34B and 7B models, respectively. For configuration (b), the speedup is 1.47x and 1.82x for the 34B and 7B models, respectively. 

For other generation tasks using the model architectures of Seamless and HSTU, we do not expect performance improvement based on the characterization results in Figure~\ref{fig:genai-model-breakdown} --- linear operations do not contribute significant runtime to end-to-end model inference. Furthermore, quantization optimization needs careful tuning, especially for production use cases of recommendation models~\cite{recquant}, thus we opt out HSTU from AutoQuant enablement.

%We apply AutoQuant to the Chameleon and Llama models, which are expected to benefit most from this optimization. We intentionally exclude the Seamless and HSTU models, as Figure~\ref{fig:genai-model-breakdown} indicates that linear operations are not bottlenecks in these models. Given that AutoQuant targets the acceleration of linear operations, Chameleon and Llama are deemed the best candidates. Additionally, production use cases for recommender system models often prohibit further quantization due to sensitivity to accuracy, which directly impacts company revenue~\cite{recquant}.

%\yejin{Should we draw graph for AutoQuant only? Or squeeze in to other graph?}
%The end-to-end inference time speedup achieved by AutoQuant is presented as blue bars in Figure~\ref{fig:torchcompile2} for the Chameleon and Llama models. We evaluate two configurations: (a) a batch size of 1 across all workloads, and (b) the maximum batch size for each workload that fits within a single NVIDIA A100 GPU, as specified in Table~\ref{tbl:tasks}. We also assess two model sizes for Chameleon and Llama, 7B and 34B, to demonstrate the varying impact of torch.compile and AutoQuant depending on compute intensity.

%For configuration (a), the models achieve a speedup of 1.53x and 2.28x for the 34B and 7B models, respectively. For configuration (b), the speedup is 1.47x and 1.82x for the 34B and 7B models, respectively. Compared to enabling torch.compile alone, AutoQuant provides additional performance improvements.

\vspace{-0.1in}
\subsection{Algorithm and NN Specific Optimizations}\label{sec:worklaodspecific}

To meet the low inference latency requirement with resource efficiency, we prioritize enabling system optimization levers that come with minimal accuracy impact --- SDPA and Flash Attention in Section~\ref{sec:sdpa}~\cite{golden2024flashattentionstable}, torch.compile and CUDA Graph in Section~\ref{sec:torch-compile}, and AutoQuant in Section~\ref{sec:autoquant}.  
To further efficiently accelerate inference, algorithm and neural network specific optimizations levers could be exploited. Here, we focus on a state-of-the-art inference optimization technique: LayerSkip~\cite{layerskip}.
%and CHAI~\cite{chai}. 
This technique is originally designed for Llama inference time and we show how it could be utilized to accelerate other multi-modal generative models. 
% Then, we compare and contrast the design space of system optimization that is horizontally enabled across the generation tasks with that of algorithmic optimization.

%In this section, we introduce two workload-specific techniques that are tailored to optimize certain type of objectives, LayerSkip~\cite{layerskip} and CHAI~\cite{chai}. Both techniques are originally proposed as optimization techniques to accelerate inference time for Llama, but we show how these techniques could be utilized to accelerate other multimodal generative models. We focus on how much performance improvement we could get from algorithmic optimization only compared to system optimization only.

\begin{figure}
     \centering
     \includegraphics[width=0.65\columnwidth]{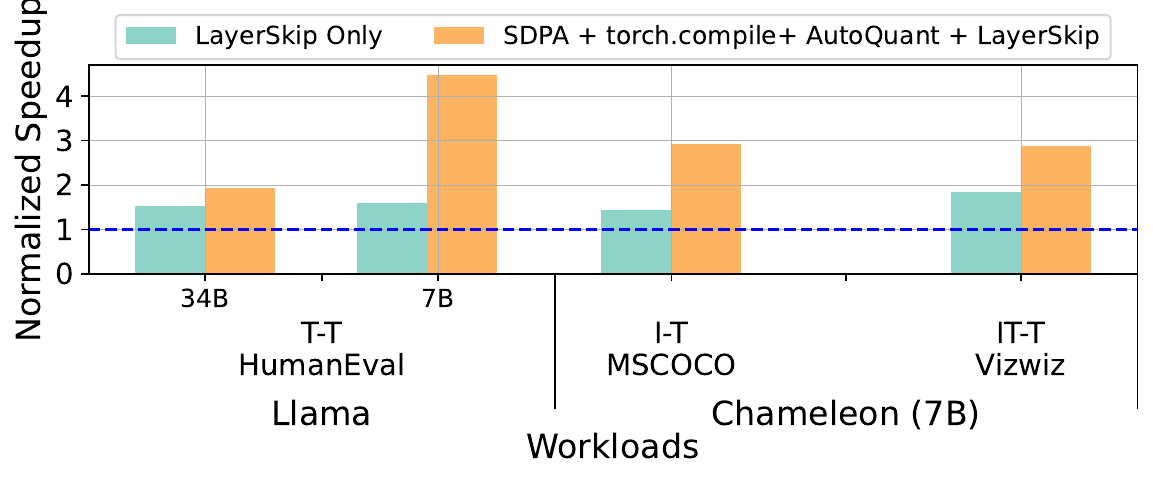}
     % \vspace{-8mm}
     \caption{End-to-end inference time speedup with LayerSkip with batch size = 1 on NVIDIA A100 GPU.}
     \label{fig:workloadspeedup}
     \vspace{-0.15in}
\end{figure}

% layer skip is proposed for single batch. supporting batch size > 1 will require change in design. optimal number to skip layer for single batch might not work for batch size > 1. it requires compromise.

LayerSkip~\cite{layerskip} is specialized to minimize single-batch inference latency of LLMs. 
It speeds up inference by generating draft tokens sequentially with fewer layers while verifying them in parallel with remaining layers. Like speculative decoding~\cite{fastinference}, parallel token verification amortizes per-token layer weight loading costs, resulting in end-to-end speedup. However, accuracy loss with exiting early is recovered by finetuning the model with a training recipe~\cite{layerskip}.
% that includes early exit loss and layer dropout~\cite{dropout}.

However, LayerSkip requires continuous model pretraining with early exit loss and layer dropout to improve early layer accuracy.
It required 64 GPUs for 50K iterations for Code Llama and Chameleon, and moreover required access to the same or similar pretraining corpus that the models were trained on. Also, LayerSkip can be beneficial only for auto-regressive decoder models such as Llama, Chameleon, but not for non auto-regressive models like HSTU. Moreover, LayerSkip requires speculative decoding implementation for its mechanism, thus a custom implementation is required.
 
% During training, layer dropout~\cite{dropout} is applied, using low dropout rates for earlier layers and higher dropout rates for later layers with an early exit loss where all layers share the same classification head. Then, during inference, the training recipe increases the accuracy of early exit at the earlier layers, without adding any auxiliary layers or modules to the model. Finally, self-speculative decoding is introduced, where a subset of earlier layers are used to generate tokens sequentially, and remaining layers are used to verify and corrects tokens in parallel, amortizing the cost of loading their weights. 
% LayerSkip provides less memory footprint than other speculative decoding approaches and benefits from shared compute and activations of the draft and verification stages. 

% \subsubsection{CHAI}
% Clustered Head Attention (CHAI)~\cite{chai} accelerates the execution time of multi-head Attention in LLMs by combining the Attention heads with a high degree of correlation for self-attention. CHAI takes advantage of the empirical observation that, in Llama, there is a high amount of redundancy across tokens which the Attention heads pay attention to. CHAI clusters the Attention heads with significant similarity patterns to reduce Llama computation complexity. It is a simple inference time optimization technique that does not require any fine-tuning. The inference latency speedup comes from reduced memory and computation requirements.

\noindent \textbf{Results -- LayerSkip.}
Figure~\ref{fig:workloadspeedup} shows the inference time performance gain from workload-specific optimizations, we choose Llama and Chameleon as our target models. We focus on batch size 1 because efficient speculative decoding for larger batch sizes require significant modification to the attention mechanism~\cite{qian2024bassbatchedattentionoptimizedspeculative, vLLMOptimizingAttention}. However, LayerSkip is an optimization technique that achieves significant inference time speedup at the cost of accuracy loss. We achieve \layerskipcodellamaseven\ and \layerskipcodellamathirty\ speedup with +2.5\% and -1.2\% accuracy impact for CodeLlama 7B and 34B model, respectively. For Chameleon 7B model, LayerSkip achieves \layerskipchameleonmscoco\ and \layerskipchameleonvizwiz\ speedup with -3.2 and -6.36 cider score loss for I-T and IT-T tasks, respectively. 
Overall, we observed the geomean \layerskipspeedup speedup only with LayerSkip. 

\noindent \textbf{Results -- Putting It Altogether.}
We further explored performance gains by enabling all cross-stack optimization techniques, system-level optimizations (SDPA, torch.compile, AutoQuant) and workload-load specific optimization (LayerSkip). This enhanced the speedup from \layerskipspeedup\ to \layerskipsystemspeedup, demonstrating the significant potential of combining techniques for optimal performance gains.

\subsection{Roofline Analysis}\label{sec:roofline}
Figure~\ref{fig:roofline} illustrates the effects of various optimization techniques on performance, as evaluated through the roofline analysis (data collected from \texttt{NSight Compute}~\cite{nsight-compute} profiling tool from NVIDIA).
For each workload, \textit{Baseline} is indicated with a circle marker where none of the optimization techniques is applied whereas \textit{Sys-Opt} is indicated with a star marker where all the optimization levers are enabled.
For Llama and Chameleon, SDPA+torch.compile+AutoQuant are enabled while SDPA+torch.co\linebreak mpile is enabled for Seamless and SDPA is enabled for HSTU.

For each workload, \textit{enabling the system-level optimization techniques increased arithmetic intensity (i.e., FLOP/memory\_traffic) and the performance (i.e., FLOP per sec)}, moving workload characteristics to the upper right part of the roofline.
In the A100 deployment case, workloads that were already memory bandwidth-bound in the baseline setup are able to reduce the memory traffic and improve overall system performance.
SDPA minimizes memory accesses during the attention mechanism by breaking down input sequence lengths into smaller tiles and perform computation within each tile independently.
torch.compile fuses operations, eliminating intermediate memory allocations and accesses.
AutoQuant effectively decreases the memory usage/traffic of each weight parameter by lowering their numerical precision.

The effect of applying each of these optimizations also depends on properties of each workload such as the underlying model architecture and input sequence length.
For instance, we observe that workloads with textual inputs (e.g., T-T, T-I) and were previously most memory bandwidth-bound were the biggest beneficiaries.
However, Seamless shows at most 10\% difference (5\% on average) between circle and star marker across four workloads (S-S, S-T, T-S, T-T). As mentioned in Section~\ref{sec:torch-compile}, applying torch.compile only to the text decoder among four primary modules in Seamless results in trivial impact to overall arithmetic intensity and the performance.

\begin{figure}
    \centering
    \includegraphics[width=0.65\columnwidth]{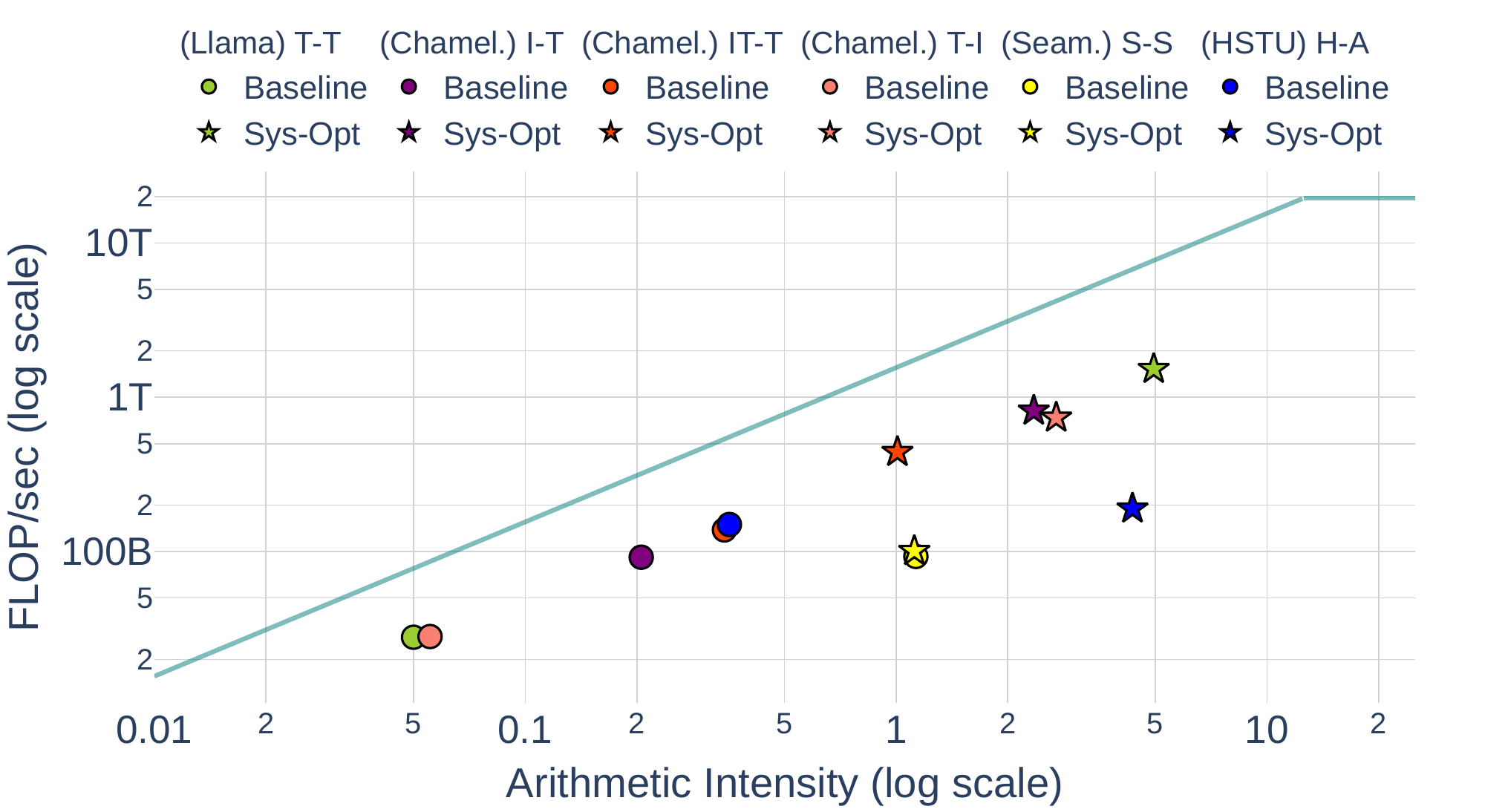}
    \caption{Roofline Analysis for Generative AI Workloads.}
    \label{fig:roofline}
    % \vspace{-0.5cm}
\end{figure}

\noindent\textbf{Beyond the Roofline Analysis} We take Llama as an example to further understand the performance implication of each optimization technique to the roofline. 
When we apply SDPA, two key effects occur: First, FLOPs count increases by 8\%.
This is because efficient attention techniques require some recomputation.
Second, memory traffic decreases by 14\% due to the optimized algorithm.
As a result, the arithmetic intensity increases.
% Applying torch.compile on top of SDPA both increased the FLOPs count and memory traffic because we adopt static KV cache, however the arithmetic intensity increased because the FLOPs count increased in bigger scale (attention has $O(N^2)$ complexity) compared to memory traffic increase.
Counterintuitively, applying torch.compile on top of SDPA both increases both FLOPs count and memory traffic due to static KV cache adoption.
Overall arithmetic intensity still increases because FLOPs count increases at a faster rate compared to memory traffic (attention has $O(N^2)$ complexity).
Memory traffic increases slightly by 1\%: the static KV cache's increased memory traffic more or less cancels out the reductions from fusing operations.
By reducing per-weight memory footprint, AutoQuant reduces the memory traffic by 3.1$\times$ on top of SDPA and torch.compile.
Applying LayerSkip on top of all system-level optimizations reduces the FLOPs count by 2.3$\times$ and the memory traffic by 2.2$\times$.

\begin{figure}
  \centering
  \includegraphics[width=0.65\columnwidth]{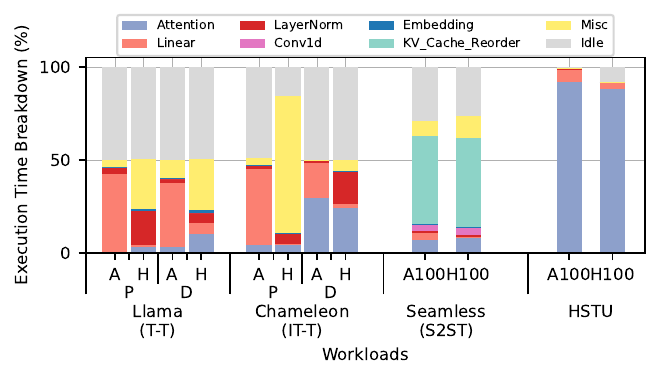}
  \vspace{-5mm}
  \caption{Operator Time Breakdown of Code Llama~\cite{codellama}, Seamless~\cite{seamless}, Chameleon~\cite{chameleon}, HSTU~\cite{hstu} on H100 GPU.}
  % \vspace{-5.25mm}
  \label{fig:h100breakdown}
\end{figure}

% \vspace{-0.6cm}

\subsection{Result Analysis over GPU Generations}\label{sec:h100}
In this section, we extend our analysis to NVIDIA H100 GPU~\cite{nvidiah100} to demonstrate how our insights and optimizations generalize across different hardware generations.
H100 (Hopper) is the newer generation GPU beyond NVIDIA A100 (Ampere), introducing improvements in both computing capability and memory subsystem, achieving about 3$\times$ higher theoretical peak FLOPS and 1.5$\times$ higher HBM bandwidth compared to A100.

By examining the same workloads on both platforms, it brings insights on how the new generation of hardware could impact the performance characteristics in diverse aspects by showing how existing bottleneck is resolved with the architectural improvements and what new optimization opportunities come up.
These cross-platform insights are crucial for both hardware architects designing future accelerators and system engineers optimizing software stack.

Figure~\ref{fig:h100breakdown} shows H100 GPU operator time breakdown compared to A100 (Figure~\ref{fig:genai-model-breakdown}). We observed two changes in performance characteristics. First, H100 demonstrates substantial improvements in computational efficiency, with \textit{Linear} operations showing the most dramatic speedup of 6.82$\times$, while \textit{Attention} operations achieve a 1.44$\times$ improvement resulting in 1.68$\times$ speedup for end-to-end \textit{baseline} runtime for batch size 1. Second, significant acceleration in Linear operation has shifted the performance bottlenecks - models previously bounded by Linear operations now show Misc or Attention operations as their primary bottlenecks.

\begin{figure}
     \centering
    \includegraphics[width=0.65\columnwidth]{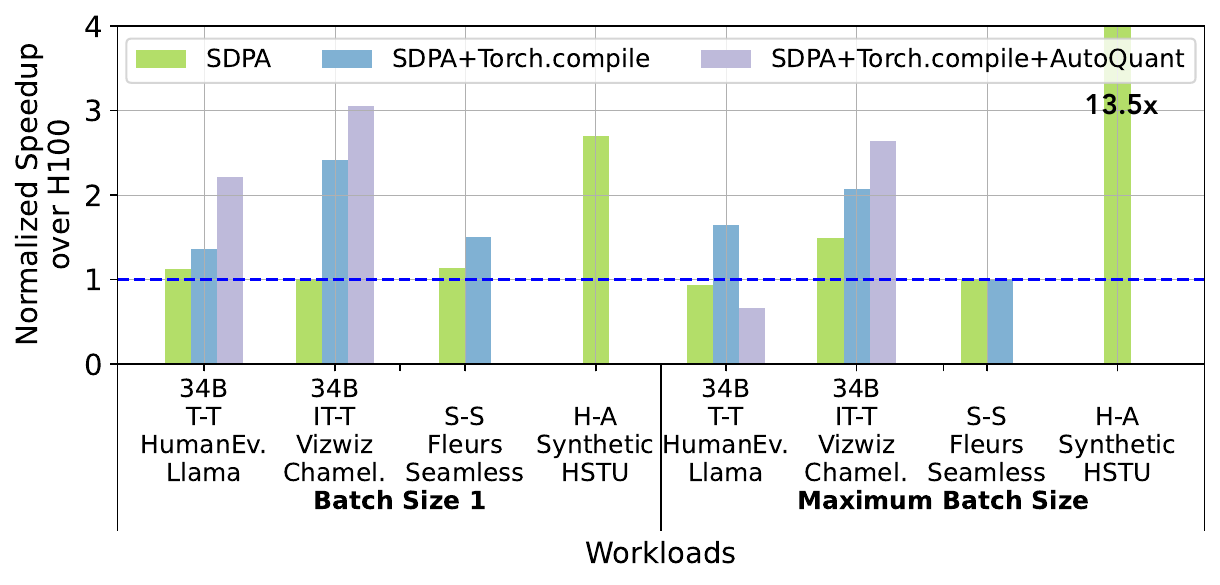}
     % \vspace{-0.9cm}
     \caption{End-to-end inference time speedup with SDPA and SDPA+torch.compile for Llama and Chameleon on H100.}
     \label{fig:h100_llama_chameleon}
     \vspace{-0.4cm}
\end{figure}

% \begin{figure}
%      \centering
%      \begin{subfigure}[b]{0.625\columnwidth}
%          \centering
%          \includegraphics[width=\textwidth]{Submission/analysis_figures_revision/sdpa/seamless_h100.pdf}
%          \caption{Seamless}
%      \end{subfigure}
%      % \hfill
%      \begin{subfigure}[b]{0.33\columnwidth}
%          \centering
%          \includegraphics[width=\textwidth]{Submission/analysis_figures_revision/sdpa/hstu.pdf}
%          \caption{HSTU}
%      \end{subfigure}
%      \caption{End-to-end inference time speedup with SDPA and SDPA+torch.compile for Seamless and HSTU on H100.}
%      \label{fig:h100_seamless_hstu}
% \end{figure}

% We get \sdpabsonenewhw and \sdpabsmaxnewhw speedup across the workloads for single and max batch size setting.

Figure~\ref{fig:h100_llama_chameleon} shows speedup with system-level optimizations. When all possible set of system-level optimization techniques are enabled for each workload, we get 2.21$\times$, 3.1$\times$, 1.5$\times$, 2.7$\times$ for Llama 34B, Chameleon 34B, Seamless (S-S) and HSTU for batch size 1 setting, respectively.
LayerSkip on top of system-level optimizations gives the final speedup of 2.21$\times$, 4.13$\times$, 3.22$\times$, 4.53$\times$ speedup for T-T of Llama 34B \& 7B, IT-T and I-T task of Chameleon 7B, respectively.

% - Chameleon - COCO : 4.53x
% - Chameleon - Vizwiz: 3.22x
% - Codellama - 34B: 2.21x
% - CodeLlama - 7B : 4.13x

% \begin{figure}
%      \centering
%      \includegraphics[width=\columnwidth]{Submission/analysis_figures/layerskip.pdf}
%      \caption{End-to-end inference time speedup with LayerSkip with batch size = 1 on NVIDIA H100 GPU.\textcolor{red}{Dummy Figure}}
%      \label{fig:h100_layerskip}
% \end{figure}

The reduced relative performance gains from optimizations on H100 compared to A100 can be attributed to enhanced baseline capabilities of H100. With architectural improvements including hardware-optimized attention mechanisms and higher memory bandwidth, it demonstrates diminishing returns in software optimization techniques as baseline hardware performance improves, despite the superior absolute performance of H100.

\section{Key Lessons and Concluding Remarks}
Generative AI technologies are reshaping the computing landscape by offering new capabilities. This paper characterizes system performance of key multimodal models from Meta: Llama, Seamless, Chameleon, and gDLRM and emphasizes distinct resource needs and performance patterns requiring tailored optimizations.
\textit{Enabling state-of-the-art system-level optimizations}, such as Flash Attention/SDPA, torch.compile, CUDA Graph and quantization, strengthen baseline performance. \textit{Enabling workload-specific optimizations} unlocks even further optimization opportunities by exploiting workload specific characteristics. We present reveals key insights for the computer architecture community:

% Our analysis reveals several key insights that are crucial for the computer architecture community:

\begin{itemize}[leftmargin=*]
    \item Multi-modal models show distinct workload patterns compared to traditional AI models. Our quantitative results demonstrate difference in latency, compute and memory requirements across modalities and tasks. For instance, we observed that T-I and IT-T tasks of Chameleon demands 1.7$\times$ more compute than HSTU, while the arithmetic intensity of HSTU is 1.25$\times$ higher.
    \item Optimization solutions must consider the whole end-to-end inference pipeline. Our research shows that focusing on isolated components may lead to suboptimal performance gains. For example, optimizing only the attention operation with SDPA gives \sdpabsmax\ improvement, and additionally optimizing linear operations with AutoQuant gives additional \autoquantoversdpabsmax\ speedup, resulting in total \autoquantovervanillabsmax\ inference speedup.
    \item While new hardware accelerators are exciting prospects, our results emphasize the importance of first exhausting state-of-the-art software optimizations. We demonstrated that enabling state-of-the-art optimization, SDPA, torch.compile, and AutoQuant, led to an \autoquantovervanillabsmax\ performance improvement across all the models, highlighting the untapped potential in existing hardware.
    \item The diversity in model architectures, modalities necessitates flexible and adaptable optimization strategies. 
    % Our work shows that a one-size-fits-all approach is insufficient, as evidenced by showing that PyTorch SDPA might not be always helpful depending on the significance of the attention operation portion of the total runtime.
    PyTorch SDPA's effectiveness varies with attention operation's proportion of total runtime, showing one-size-fits-all approach is insufficient.
    % \item Flexibility and adaptability should be key considerations in hardware design for generative AI tasks. Our research highlights the diverse computational patterns and requirements across different generative AI models and tasks with different optimization techniques. Thus it is important to have reconfigurable hardware design that can flexibly and efficiently handle these variation patterns. Additionally, while our study didn't extensively cover distributed inference, it's crucial to address the growing network demands of large-scale generative AI models by increasing on-chip memory or enhancing inter/intra host communication
    \item Hardware design for generative AI tasks should prioritize flexibility and adaptability to accommodate diverse computational patterns and requirements across models, tasks and optimization knobs. Reconfigurable hardware design is essential to efficiently handle these variations. Addressing growing network demands through increased on-chip memory or enhanced inter/intra host communication is crucial for large-scale generative models.
\end{itemize}

We hope this work provides deeper understanding and insights on the landscape of generative AI technologies and cross-stack system optimization solutions. 
Focusing on optimizing fundamental components and considering unique input modalities of the key generative AI technologies are the key to efficiently accelerate model inference. The findings and methodologies in this paper enhance our understanding of generative AI system performance and set the stage for future innovations, leading to more efficient and scalable AI systems. As the field of generative AI continues to evolve rapidly, we believe that the computer architecture community has a crucial role to play in shaping the next generation of efficient, high-performance AI systems.

\section{Acknowledgment}
This work is an outcome of the extensive collaborations with many teams: Chameleon, Seamless, and HSTU. We are thankful for the valuable insights, numerous discussions, and refinement on the multimodal models. We would also like to thank the PyTorch and the xFormers teams, especially their inputs on ML system optimization.

\clearpage
\newpage
\bibliographystyle{assets/plainnat}
\bibliography{paper}

\clearpage
\newpage
% \beginappendix

% \section{First appendix}

\end{document}